\documentclass{article}

\PassOptionsToPackage{numbers, compress}{natbib}

\usepackage[final]{neurips_2022}

\usepackage{enumitem}
\usepackage{url}
\usepackage{color}
\usepackage{soul}
\usepackage[T1]{fontenc}
\usepackage{booktabs} 
\usepackage{mathtools}
\usepackage{amssymb}
\usepackage{threeparttable}
\usepackage{placeins}
\usepackage{caption}
\usepackage[]{subcaption}
\usepackage{amsfonts}       
\usepackage{nicefrac}       
\usepackage{microtype}      
\usepackage{xcolor}         
\usepackage{graphicx}

\usepackage{hyperref}
\hypersetup{
    colorlinks=true,
    linkcolor=blue!70!green,
    filecolor=magenta,      
    urlcolor=cyan,
    citecolor=blue!70!green,
}

\definecolor{lightgrey}{rgb}{0.925, 0.925, 0.925}
\definecolor{orange}{rgb}{0.925, 0.5, 0.0}

\sethlcolor{lightgrey}

\newcommand{\comb}[1]{{\small\texttt{\hl{[#1]}}}}

\title{Lost in Latent Space: \\ Examining failures of disentangled models at combinatorial generalisation}

\author{%
Milton L. Montero$^{1,2}$\thanks{Corresponding author} \quad Jeffrey S. Bowers$^1$ \quad Rui Pont Costa$^2$ \\ 
\textbf{Casimir J.H. Ludwig}$^1$ \quad \textbf{Gaurav Malhotra}$^1$ \\
$^1$School of Psychological Science \quad $^2$Department of Compute Science \\
University of Bristol \\
\texttt{\{m.lleramontero,j.bowers,rui.costa,c.ludwig,gaurav.malhotra\}@bristol.ac.uk}\\
}

\begin{document}

\maketitle
\vspace{-0.4cm}
\begin{abstract}
Recent research has shown that generative models with highly disentangled representations fail to generalise to unseen combination of generative factor values. These findings contradict earlier research which showed improved performance in out-of-training distribution settings when compared to entangled representations. Additionally, it is not clear if the reported failures are due to (a) encoders failing to map novel combinations to the proper regions of the latent space, or (b) novel combinations being mapped correctly but the decoder being unable to render the correct output for the unseen combinations. We investigate these alternatives by testing several models on a range of datasets and training settings. We find that (i) when models fail, their encoders also fail to map unseen combinations to correct regions of the latent space and (ii) when models succeed, it is either because the test conditions do not exclude enough examples, or because excluded cases involve combinations of object properties with its shape. We argue that to generalise properly, models not only need to capture factors of variation, but also understand how to invert the process that causes the visual input.

\end{abstract}

\section{Introduction}\label{intro}
\vspace{-0.2cm}

Disentangled representations extract factors of variations from data, and learning them has been the focus of much research in recent years \citep{higgins2017}. Several approaches have been proposed to induce disentanglement, including latent space penalization \citep{higgins2017,burgess_understanding_2018,kim_disentangling_2019}, different training regimes \citep{locatello2020weakly,lin2020infogan}, architectures \citep{watters_spatial_2019}, data-driven inductive biases \citep{klindt2020towards} and model selection methods \citep{duan_unsupervised_2020}. These have produced more interpretable representations \citep{higgins_darla_2018} that improve sample efficiency and learning for downstream models \citep{higgins_scan_2018,van_steenkiste_are_nodate,duan_unsupervised_2020}.

Importantly, researchers have also hypothesized that disentangled representations could provide a way of improving generalisation performance by enabling the discovery of causal variables in data \citep{scholkopf2021toward} or by capturing its compositional structure \citep{duan_unsupervised_2020}. This claim is especially interesting as it allows machine learning systems to emulate a key property of human intelligence -- the ability to generalise to unseen combinations of known elements. For example, if a model has learned to generate red triangles and blue squares, then the model should also be able to correctly generate blue triangles and red squares. This property, which we refer to as \emph{combinatorial generalisation}, gives humans the ability to make ``infinite use of finite means'' \citep{von1999humboldt,chomsky2014aspects,smolensky_connectionism_1988,mccoy2021infinite} and has been termed ``a top priority for AI to achieve human-like abilities'' \citep{battaglia2018relational}. Indeed, several authors have reported that unsupervised models that are better at disentangling generative factors are also better at some forms of combinatorial generalisation \citep{higgins2017,higgins_scan_2018,watters_spatial_2019}
\texttt{\begin{figure*}[t!]
\centering
\includegraphics[width=0.8\textwidth]{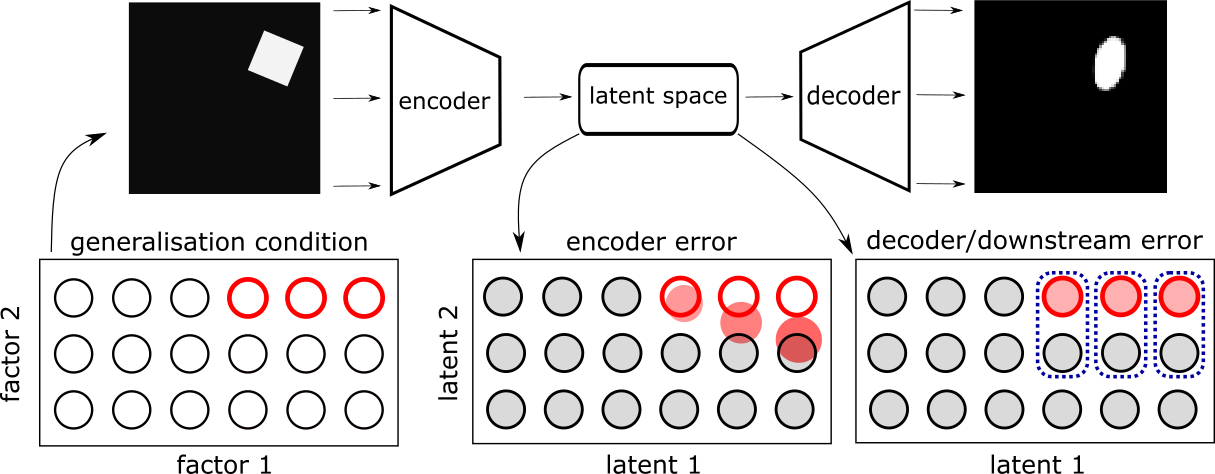}
\caption{\textbf{Possible problems for combinatorial generalisation}. \emph{Left}: visualisation of how a combinatorial generalisation condition can be defined for a dataset containing two generative factors. Black and red circles denote training and test examples, respectively. In the Middle and Right panels, the outline of each circle shows the target position of the representation in latent space, while a shaded circle shows the position at which the latent representation is projected by the encoder. \emph{Middle}: first type of combinatorial generalisation error, encoder error. The encoder projects unseen inputs to different parts of the latent space than what was expected based on their generative factor values (shown as shaded circles falling outside their target outlines) \emph{Right}: second type of combinatorial generalisation error, decoder error. Observed representations (shaded circles), are mapped to the correct position in latent space (circle outlines), but a decoder/downstream process mixes-up the black and red representations (blue dashed lines).}
\label{fig:schematic}
\vspace{-0.6cm}
\end{figure*}}

However, two recent studies have reported evidence that contradicts this hypothesis. \citet{montero2020role} considered datasets where inputs varied along several dimensions (generative factors) and divided generalisation conditions into different types. They found that VAEs with highly disentangled latent representations succeeded at the easiest generalisation conditions where only one combination of all generative factors was excluded (a condition they termed \emph{recombination-to-element}), but failed at more challenging conditions, where all combinations of a subset of generative factors were excluded (termed \emph{recombination-to-range} by \citet{montero2020role}). In another second study, \citet{schott2021visual} tested 17 learning approaches, expanding the analysis to other paradigms, architectures and task structures. They observed that models showed moderate success at generalising on some artificial datasets, such as 3DShapes. However, their ability to generalise dropped significantly on more realistic datasets and in more challenging generalisation conditions.

These contradictory results could be down to a number of differences between studies that report successes and those that report failures. For example, there were qualitative differences in the datasets that were tested and how the test conditions were set up. This is indeed what \citet{montero2020role} conclude, writing that ``it is not clear what exactly was excluded while training [previous models]'', suggesting that previously observed successes may have been on the simplest generalisation condition where only very few combinations were left out. But this still leaves open the question of why models that achieve high degree of disentanglement nevertheless fail at combinatorial generalisation under more challenging conditions.

We see two possible reasons why this could happen. One possibility is that models correctly infer the values of latent variables (Fig.~\ref{fig:schematic}, left) for novel combinations of generative factors, but the decoder fails to map these unseen latent values to the (output) image space -- \emph{decoder error} (Fig.~\ref{fig:schematic}, right). Some evidence supporting this hypothesis was observed by \citet{watters_spatial_2019}, who looked at not only the output image reconstructions, but also at latent representations in a simple reconstruction task. \citet{watters_spatial_2019} observed that, in their task, models that have more disentangled latent representations do indeed map unseen combinations of generative factors to the correct values in the latent space. But they only tested a very simple dataset, involving only two generative factors. \citet{montero2020role} showed that models can sometimes succeed to generalise in simple settings where they can solve the combinatorial generalisation problem through interpolation, but struggle at more challenging settings where larger number of combinations are excluded from the training set.

\begin{figure*}[h!]
    \begin{center}
        \includegraphics[width=\textwidth]{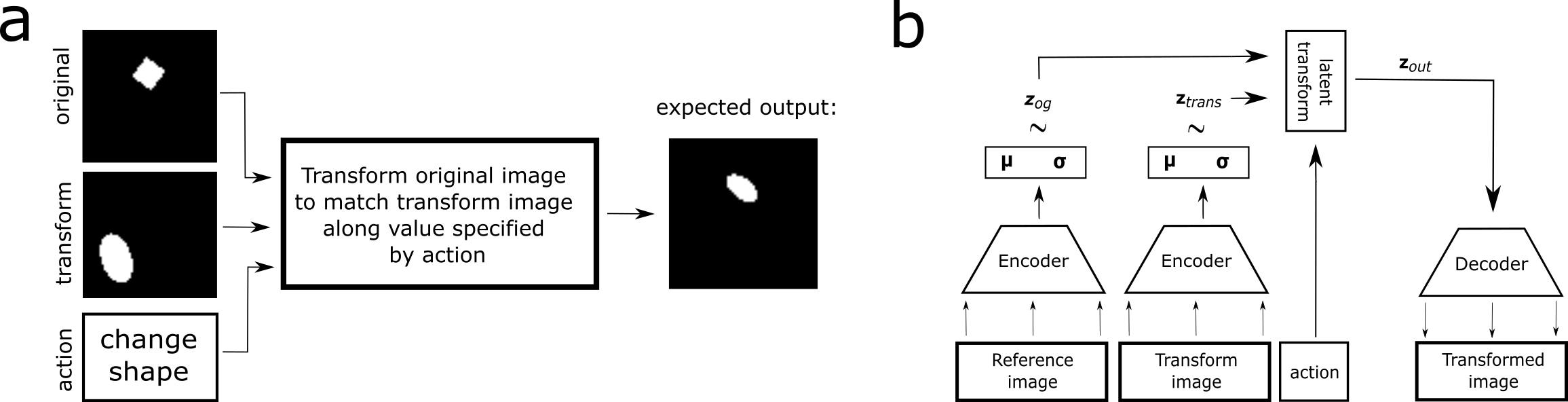}
    \end{center}
    \caption{ (a) \textbf{Image composition task}. An example of the composition task. In this case, the \comb{shape} of the output must match the \textit{transform} image and the rest of the values (\comb{position, orientation}) must match the \textit{original} image. (b) \textbf{Model architecture}. The model uses the same encoder to represent both images. Then a transform takes these latent representation and combines them to produce a transformed representation, which is used to reconstruct the output image. Reproduced from Montero et al. with permission from the authors.}
    \label{fig:task-model-2}
    \vspace{-0.5cm}
\end{figure*}

Another possibility is that current models fail at harder forms of combinatorial generalisation (such as recombination-to-range) due to an encoding failure -- \emph{encoder error}. That is, the encoder fails to map these harder unseen combinations of generative factors to the correct values of variables in the latent space (Fig.~\ref{fig:schematic}, middle). If this were the case, then it reflects a more fundamental limitation of current models, showing that even though models can map observed combinations to correct values of latent representations, they do not understand how generative factors are combined. So when novel combinations are presented, models cannot infer the values of generative factors that led to the observed data.

In this work we explore these issues by examining the latent space of generative models. Extending the findings of \citet{montero2020role} and \citet{schott2021visual}, we show that failure to generalise in the output space is accompanied by failure to generalise in latent space across several datasets. We show that these results hold for a broad range of decoder architectures (the standard \textit{deconvolution decoder} and the \textit{spatial broadcast decoder} \citep{watters_spatial_2019}), loss functions (\textit{$\beta$-VAE} and \textit{WAE}) and task settings. Finally, by looking at the latent space we discover that, in addition to the difficulty of the generalisation task, a crucial condition for failure of combinatorial generalisation is the way in which generative factors are combined. These results not only challenge research that argues that disentanglement leads to better generalisation \citep{higgins_scan_2018,watters_spatial_2019}, but also shows that the failure to generalise is a lot more entrenched in current models than suspected by \citet{montero2020role} and \citet{schott2021visual}.

\section{Testing generalisation in latent representations}\label{sec:composition}
\vspace{-0.2cm}

In this section we first describe a carefully designed semi-supervised task that enables us to achieve latent representations that are highly disentangled. This is crucial because standard disentanglement learning objectives, which rely on minimizing total correlation \citep{higgins2017,burgess_understanding_2018,kim_disentangling_2019}, struggle to reliably achieve a high degree of disentanglement, which makes it difficult to assess whether test data are projected to the expected position in the latent space (see Figure~\ref{fig:schematic}). Secondly, we describe our choice of corresponding model architectures as well as loss functions. For our results to be broadly applicable to existing research, we chose standard encoder and decoder architectures as well as a more sophisticated approach designed to induce a high degree of disentanglement \citep{watters_spatial_2019}. Lastly, we describe some challenging test conditions for combinatorial generalisation on well known datasets.

\subsection{Experimental Setup}\label{subsec:expset}

\subsubsection{Task}

One of the main barriers to achieving disentanglement in standard unsupervised training is the non-identifiability of the models when using iid data \citep{hyvarinen2016unsupervised,klindt2020towards}. In short, this is because there are infinitely many linear combinations of the underlying generative factors that produce a valid basis on which they can be represented.

To get around this problem we used the composition task developed by \citet{montero2020role} (see Appendix~\ref{app:task} for further discussion). This task takes two images and a query vector as inputs (see Figure~\ref{fig:task-model-2}(a)). The goal of the task is to output an image that combines the two images based on an \emph{action} in the query vector:

\begin{align*}
    \textbf{input} &= \mathbf{x_{og}}, \; \mathbf{x_{trans}}, \; \mathbf{q} \\
    \textbf{output} &= \mathbf{x_{out}}
\end{align*}
where $\mathbf{x_{og}}$ and $\mathbf{x_{trans}}$ are the two input images, $\mathbf{x_{out}}$ is the output image, and $\mathbf{q}$ is the query vector. Following \citeauthor{montero2020role}, we used actions that involve replacing one of the properties (generative factors) of $\mathbf{x_{og}}$ with a property of $\mathbf{x_{trans}}$, based on the value of $\mathbf{q}$. This vector was a one-hot encoding of the generative factor that was required to be changed.

By combining a semi-supervised component (the query vector plus image reconstruction) with sparse transitions (only one generative factor needed to be modified at a time), the composition task provided a strong inductive bias towards high levels of disentanglement, as has been shown in previous work that exploits these approaches \citep{locatello2020weakly,lin2020infogan,klindt2020towards}.

\subsubsection{Models}

The model architecture used to solve the composition task is shown in Figure~\ref{fig:task-model-2}(b). It involves two encoders that map each image to a latent space. The latent vectors are then ``composed'' (see below), based on the query (action) vector. This composed vector is mapped from the latent space into the (output) image space using a decoder. Encoders and decoders for generative models match the ones used in past work \citep{higgins2017,kim_disentangling_2019} for both \texttt{dSprites} and \texttt{3DShapes} datasets. For \texttt{MPI3D} we added another 2 more layers, which we found was necessary for models to learn the task. For the Circles dataset we used the encoder defined in Watters et al, along with the corresponding Spatial Broadcast Decoder architecture. Latent representations for the generative models are standard diagonal Gaussians with input dependent means and variances, to which we applied both the Variational and Wasserstein loss \citep{kingma_auto-encoding_2013,tolstikhin2017wasserstein}.

We can define the composition operation that combines both latent representations as:
\begin{equation*}
\mathbf{z_{out} = z_{og} \odot (1 - c) +  z_{trans} \odot c}
\end{equation*}
where $\mathbf{z_{out}}$, $\mathbf{z_{og}}$ and $\mathbf{z_{trans}}$ are the latent representations corresponding to the output, original, and transform images and $\odot$ is the element-wise product between vectors. The variable $\mathbf{c}$ is the interpolation vector of coefficients. We defined two ways of computing $\mathbf{c}$: (i) a learned interpolation function computed as $\sigma(W \cdot [\mathbf{z_{og};z_{trans};q}] + \mathbf{b})$ with parameters $W$ and $\mathbf{b}$ where $[.;.]$ represents the concatenation of vectors; and (ii) a fixed interpolation function $\mathbf{c}=\text{padded}(\mathbf{q})$ which just pads $\mathbf{q}$ with zeros to match the vector length of the latent representation (see Appendix~\ref{app:models} for more details).

To facilitate our goal of testing combinatorial generalisation, we augmented the target output by requiring the model to reconstruct the input images as well. Thus the trained encoder and decoder form a valid standard autoencoder. We can then probe the image reconstructions and the latent representations for unseen combinations to check if the models generalise. An illustration of this model can be found in Figure~\ref{fig:task-model-2}.

\subsubsection{Datasets}

We tested our models on three standard datasets: \texttt{dSprites} (a sythesised dataset of 2D shapes procedurally generated from 5 independent generative factors \citep{matthey_dsprites_2017}), \texttt{3DShapes} (another synthesised dataset consisting of 3D shapes procedurally generated from 6 ground truth generative factors \citep{3dshapes18}) and \texttt{MPI3D} (a collection of four different datasets consisting of synthetic as well as real-world images procedurally generated from 7 generative factors \citep{gondal2019}). These are the most popular datasets used in the disentanglement literature that also provide ground truth generative factor values.
\begin{figure*}[ht]
\includegraphics[width=\textwidth]{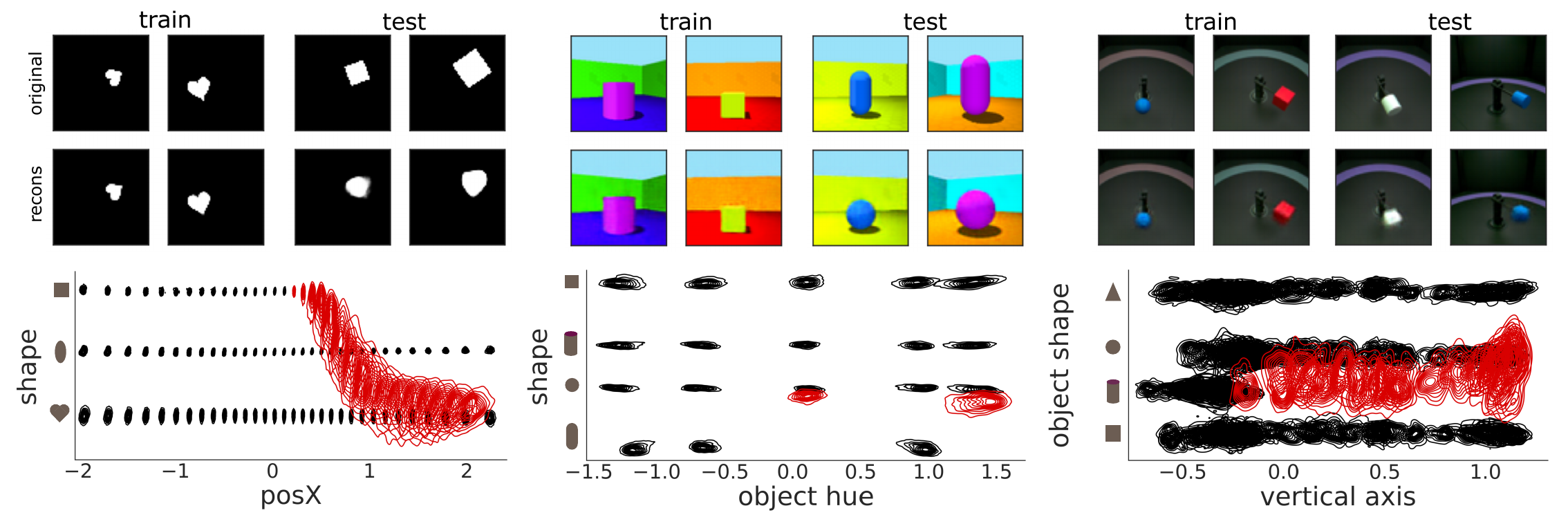}
\caption{\textbf{Generalisation in latent space} \textit{Top}: Typical examples of reconstructions for training and test images for models that learn highly disentangled representations. \textit{Bottom}: Visualisation of the latent space of models trained on three datasets -- \texttt{dSprites} (left), \texttt{3DShapes} (middle) and \texttt{MPI3D} (right). Each panel shows contours from the joint distribution of two latent variables that best predict the corresponding generative factors. In all cases, the red distributions indicate test data and the black ones indicate training data. Note that the latent values in this figure are \emph{not} necessarily the same as the values of generative factors excluded. This is because every model ends up with a different internal representation based on its initialisation and sequence of training trials. However, in each case, we observed that these internal representations were highly structured when the model showed a high degree of disentanglement.}
\label{fig:results-comp}
\vspace{-0.5cm}
\end{figure*}
We tested combinatorial generalisation by systematically excluding combinations of values of generative factors from each dataset. Here we focus on the conditions described as \emph{recombination-to-range} by \citeauthor{montero2020role}. These are the most interesting conditions where, according to \citeauthor{montero2020role}, one would expect a model that learns disentangled representations to succeed at combinatorial generalisation, but tested models typically fail. Consider a dataset with, say, four generative factors \comb{$g_1$, $g_2$, $g_3$, $g_4$} where all $g_i \in [0,1]$. The \emph{recombination-to-range} condition creates a training/test split where all examples with combinations of a subset of generative factors are excluded from the training set and added to the test set. Thus, an example of a dataset that tests recombination-to-range may consist of a training set where all combinations where \comb{$g_1 > 0.5$, $g_2 > 0.5$} have been excluded from the training set and added to the test set. Note that the model trained on such a datasets would come across a number of examples where \comb{$g_1 > 0.5$} and also examples where \comb{$g_2 > 0.5$}, but never be trained on an example where both these conditions are true simultaneously. This method was used to create training / test sets for each of the datsets in the following manner:
\begin{itemize}[noitemsep,topsep=0pt,leftmargin=10pt]
    \item \texttt{\textbf{dSprites}}: All images such that \comb{shape=square, posX$>0.5$} were excluded from the training set. Squares never appear on the right side of the image, but do appear on the left and other shapes (hearts, ellipsis) are observed on the right as well.
    \item \texttt{\textbf{3DShapes}}: All images such that \comb{shape=pill, object-hue=$>0.5$} were excluded from the training set. Thus, pills colored as any of the colors in the second half of the HSV spectrum did not appear in the training set. These colors (shades of blue, purple, etc) were observed on the other shapes, and the pill was observed with other colors such as red and orange. 
    \item \texttt{\textbf{MPI3D}}: All images such that \comb{shape=cylinder, vertical axis$>0.5$} were excluded from the training set. We also excluded all images where \comb{shape=cone} or \comb{shape=hexagonal} as these shapes are very similar to the pyramid and cylinder, respectively and make it hard to access reconstruction accuracy.
\end{itemize}
\subsubsection{Measuring Disentanglement}\label{sec:disent}
We used the DCI metric from \citet{eastwood_framework_2018} to measure the degree of disentanglement (see Appendix~\ref{app:disent-def}). This metric provides a set of regression weights between ground truth factors and latent variables. We selected the latent variable corresponding to each ground truth factor using a matching algorithm based on these weights. Crucially, for highly disentangled models any combination of generative factors will have a corresponding combination of latent variables of the same cardinality. This allows us to create visualisations of the latent representations analogous to the ones in Figure~\ref{fig:schematic}.

\subsection{Results}
We observed that most models trained on the composition task managed to achieve a reasonably high degree of disentanglement, and successfully reconstructed images in the training data (see Appendix~\ref{app:scores} and~\ref{app:recons} for quantitative results). Here we present results of typical models that achieved a very high degree of disentanglement (with alignment scores typically $>0.95$, see Appendix~\ref{app:scores} for details). Figure~\ref{fig:results-comp} shows reconstructions as well latent space representations of three typical models trained on the \texttt{dSprites}, \texttt{3DShapes} and \texttt{MPI3D} datasets, respectively.

Replicating results of \citet{montero2020role}, we observed that models showed poor generalisation to unseen combinations for all three datasets (see reconstruction of test images in Figure~\ref{fig:results-comp}). For \texttt{dSprites}, the models typically produced images in which the target location is correct but the shape is not (e.g., replacing the square with an ellipse). Similarly, with \texttt{3DShapes} and \texttt{MPI3D}, models typically reconstructed shapes with the correct colour hue (\texttt{3DShapes}) or vertical location (\texttt{MPI3D}), but made an error in the shape of the target object (e.g., replacing the pill (\texttt{3DShapes}) or cylinder (\texttt{MPI3D}) with a sphere(\texttt{3DShapes}) or pyramid (\texttt{MPI3D}).

\begin{figure*}
\centering
\includegraphics[width=0.75\textwidth]{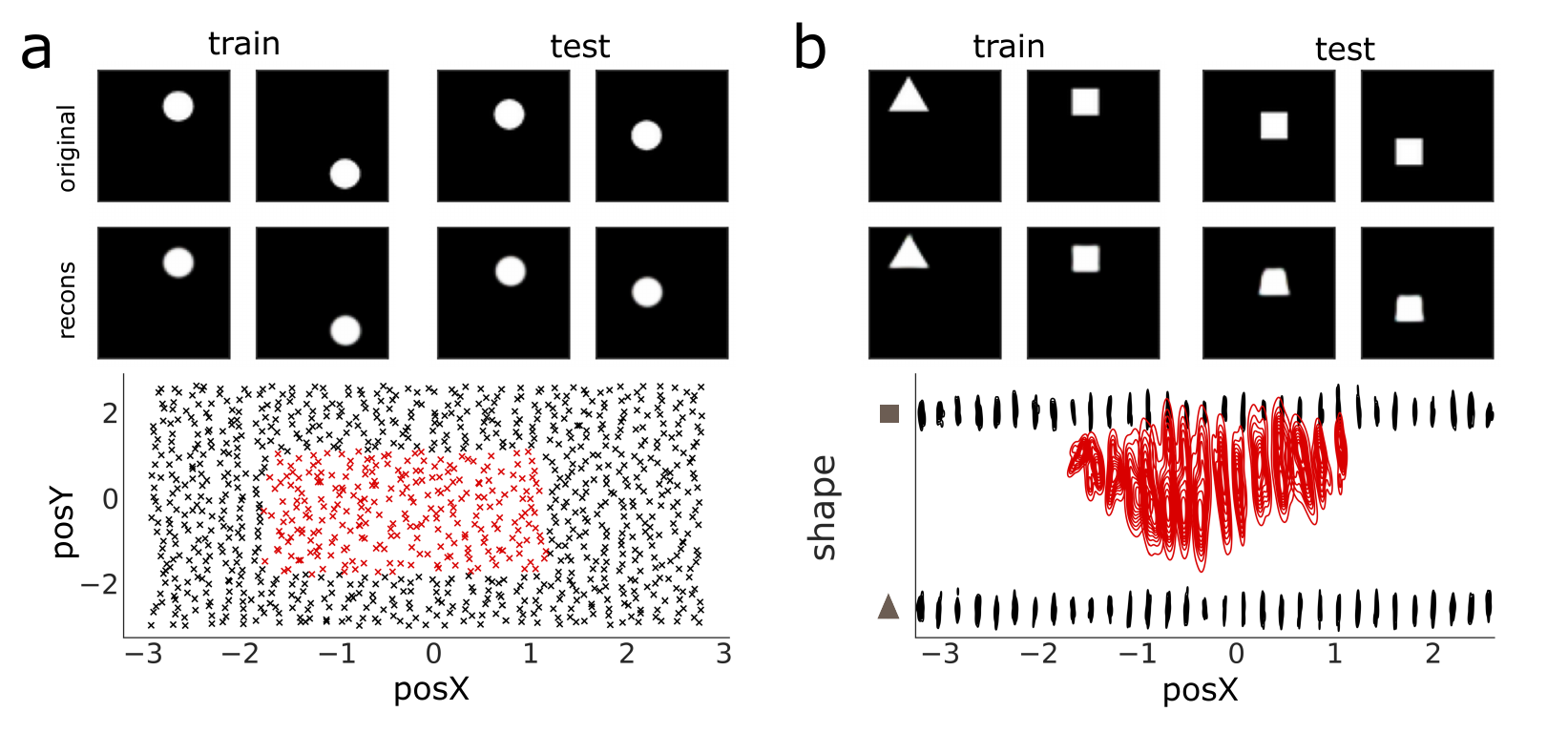}
\caption{\textbf{Latent space induced by SBD} Visualisation of the latent space for models that use a Spatial Broadcast Decoder as described in Watters et al. Both visualisations correspond to a condition where combinations in the middle of the image are removed. (a) The result when this is done for the Circles dataset which uses only one shape and where the training set excludes all combinations \comb{shape=circle, $0.35<$~posX~$<0.65$, $0.35<$~posY~$<0.65$} (b) The result for the a similar condition for the Simple dataset which uses two shapes and excludes all combinations \comb{shape=square, $0.35<$~posX~$<0.65$, $0.35<$~posY~$<0.65$}.}
\label{fig:results-sbd}
\vspace{-0.4cm}
\end{figure*}

Next, we looked at the latent representations of each of these cases by selecting a latent variable that showed highest correlation with each generative factor as described in Section~\ref{sec:disent}. Figure~\ref{fig:results-comp} plots the distributions of the value of each latent variable for every combination of values of the relevant generative factors (see Appendix~\ref{app:disent-def} for details). Visualising these probability distributions for the combination of generative factors seen during training (in black) confirmed that the models learned highly disentangled representations (note the low variance of the probability distributions for trained combinations, especially for \texttt{dSprites} and \texttt{3DShapes}).

Crucially, we observed that the encoder failed to map the generative factors for the test combinations to the correct location in the latent space (red distributions). This was particularly true for \texttt{dSprites} and \texttt{3DShapes} dataset: note how the mean of the probability distributions for the left out combinations (\comb{shape=square, posX$>0.5$} for \texttt{dSprites} and \comb{shape=pill, object-hue=[blue, purple]}) are shifted to overlap the latent representations seen during training. The shifts in the  mean of distributions were less acute for \texttt{MPI3D}, but we nevertheless observed an increase in the variability of test distributions and a consistent shift in the location for values of generative factors that were far from the trained values.

In summary, we made three key observations in these experiments. First, models replicated the negative results reported previously: i.e., a degradation in reconstruction performance in the challenging combinatorial generalisation conditions. Second, by visualising the latent space, we observed that failures of generalisation coincided with poor latent representations, showing that failures of generalisation are not entirely due to decoder errors. Third, the failures in latent representations as well as output reconstructions showed that different generative factors have important qualitative differences, with models making larger errors in generating the correct shape for unseen combinations than in reproducing the position, scale or color (see Appendix~\ref{app:recons}).

\subsection{Exploring alternative hypotheses}\label{sec:alt-architectures}

In the experiments above, we used the deconvolution network as the decoder, which is the standard decoder used in VAEs \citep{higgins2017}. However, some studies have shown that replacing the deconvolution network with a different architecture helps the model learn representations that are more disentangled. One such architecture is the Spatial Broadcast Decoder (SBD from hereon) developed by \citet{watters_spatial_2019}, who showed that such a decoder not only helped in learning representations that were highly disentangled, but also helped in solving the problem of combinatorial generalisation.

In our next set of experiments, we tested whether using the SBD succeeds at combinatorial generalisation in more challenging settings, akin to the ones we tested above. We first replicated the results for both conditions tested by \citet{watters_spatial_2019}, using our composition task on the Circles dataset and replacing the deconvolution network with the SBD. Figure~\ref{fig:results-sbd} (panel A) shows the results for the first condition, where the training set excluded images with circles presented in the middle of the canvas (see Appendix~\ref{app:recons} for results on the second condition). As we can see from this figure, the latent space learned by the  model was indeed highly structured and, crucially, the model mapped the latent values for the left out combinations (red crosses) to the correct region of the latent space.

Next, we wanted to check how these results scaled when we made the generalisation test more challenging, excluding a combination of values for a range of variables (the \emph{recombination-to-range} condition developed by \citet{montero2020role}). We can do this with the Circles dataset by adding a third generative factor and the simplest way of doing this is by letting the shape take one of several values, instead of being a circle for all examples.

\begin{figure*}
\includegraphics[width=\textwidth]{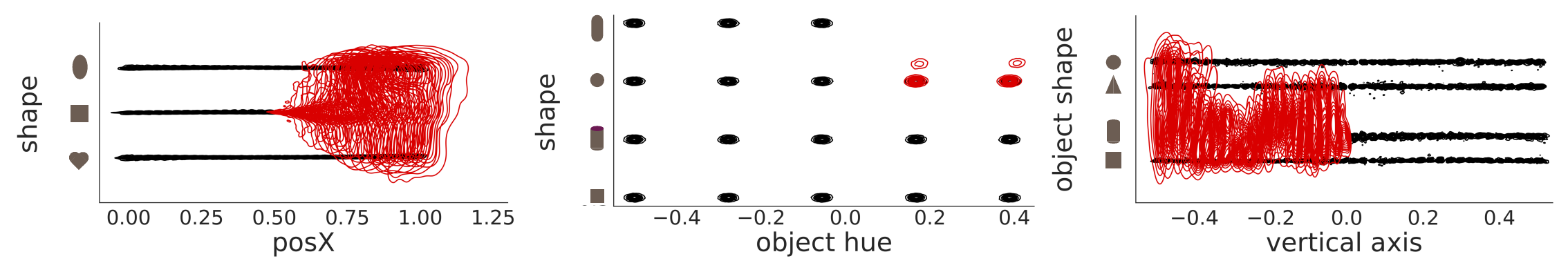}
\caption{\textbf{Replacing decoder with ground-truth} Visualisation of the latent space for models trained to predict the correct generative factor values for each image. The visualizations correspond to the same test conditions as the ones presented in Figure~\ref{fig:results-comp}: (a) \texttt{dSprites}, where all combinations where \comb{shape=square, posX~$>0.5$} were excluded from the training set (b) \texttt{3DShapes}, where all combinations where \comb{shape=pill, object-hue~$>0.5$} were excluded, and (c) \texttt{MPI3D}, where all combinations where \comb{shape=cylinder, vertical-axis~$>0.5$} were excluded (note that, in this case, \comb{vertical-axis~$>0.5$} corresponds to the corresponding latent variable value $<0$).}
\label{fig:superv-results}
\vspace{-0.5cm}
\end{figure*}

We tested this by constructing a new dataset called Simple. Like the Circles dataset, this dataset consisted of images that had a shape located at various (x and y) positions on the canvas. Unlike the Circles dataset, this shape could be either a triangle or a square (we chose these shapes as they are different enough for models to not confuse them by accident). We then tested the two conditions similar to the conditions tested by \citet{watters_spatial_2019}, excluding all combinations where one of the shapes was presented in the middle (first condition) or the top-right (second condition) of the canvas.


The results for the first condition are shown in panel~B of Figure~\ref{fig:results-sbd} (see Appendix for results of the second condition). Like the results for the first set of experiments above, we have plotted the probability distribution of combinations of values of the the latent variables (here, shape and posX) for all values of the third latent variable (here, posY). Similar to our earlier results, we observed that even the model using the SBD failed to map the unseen (test) combinations to the correct position in the latent space (compare the mean and variability of the red distributions compared to the black ones). Correspondingly, we also observed that models failed to correctly reconstruct output images for the test conditions, even when they had no problem reconstructing these images for combination of generative factors seen during training (See Appendix~\ref{app:recons}).

\section{Understanding the role of the encoder}
\vspace{-0.3cm}

We have shown that errors in generalisation occur in latent space as well in the reconstructions for generative models. To conclude this analysis we show that the problem does not lie in the generative nature of the task by testing the encoders on a latent prediction task, where the target is not the ground truth factor values. We used the same datasets and generalisation conditions as before.

\subsection{Experimental setup}

We trained models on a prediction task where the model was given a single image as the input and the target was to output the ground truth generative factor values. This ensures that the resulting latent representation will be completely disentangled, as the model must output each of the factor values separately. We used the mean squared error of the output and target vectors as a learning signal. All other parameters remain same as the composition task. We used the same architectures for the feed-forward models as the encoders in the corresponding datasets above.

We evaluated the models using the $R^2$ metric as used in \citet{schott2021visual} to check whether the models had solved the task. We also use the same method to visualise the results as described above, plotting the joint probability distributions for the combination of generative factors along which the model was tested for generalisation. In this case, it is straightforward to do so as the relevant output dimensions shared the same index as the corresponding generative factor.

Because a supervised learning setting might not give such a rich signal to the model, we also tested a different scheme that still tests the encoder. In this alternative setting, we first train a complete generative model on the full data set to a high degree of disentanglement. We then freeze the decoder and \emph{retrain the encoder} on the generalisation conditions we are interested in (see Figure~\ref{fig:frankenstein-schema} in Appendix~\ref{app:frankenstein}). Thus, the encoder has to reproduce the latent representations that the decoder is expecting in order to properly reconstruct unseen samples. Simultaneously, errors can only be a product of failures by the encoder, as the decoder is completely disentangled during its training.

\subsection{Results}

The results for the three dataset are shown in Figure~\ref{fig:superv-results}. We observed a very similar pattern for this task as the semi-supervised task above. For all three datasets, the probability distributions of latent values for the left out combinations (in red) showed a much higher variance and were shifted towards combinations of latent values that had been experienced during training. Thus, even when the encoders were trained to recognise perfectly disentangled generative factors, they failed to generalise to combinations that were not experienced during training. For the frozen decoder setting, we include results in Appendix~\ref{app:frankenstein}, showing that the failures occur there as well.

\section{Explaining contradictory findings}
\vspace{-0.2cm}

How can we reconcile the failures of generalisation observed above with with past studies that showed successful combinatorial generalisation on some datasets and conditions? One response is that the generalisation conditions we test are more challenging than the ones tested in previous studies because they exclude more data from the training set. However, it is also possible that the success of combinatorial generalisation is not only determined by how many combinations are excluded from the test condition, but also by which combination of generative factors are tested.

All failures observed above were for cases where the generative factor `shape' is combined with another generative factor (position / color). It is possible that this was because shape combines with other factors in a qualitatively different manner than, say, the floor hue combines with the wall hue in \texttt{3DShapes}. In order to completely disentangle shape from, say, position the model must come up with a representation of shape that is invariant to a change in position. And it must do so by learning from training examples where shape and position jointly determine the same part of the image.

\begin{figure}[htbp]
    \centering
    \includegraphics[width=\textwidth]{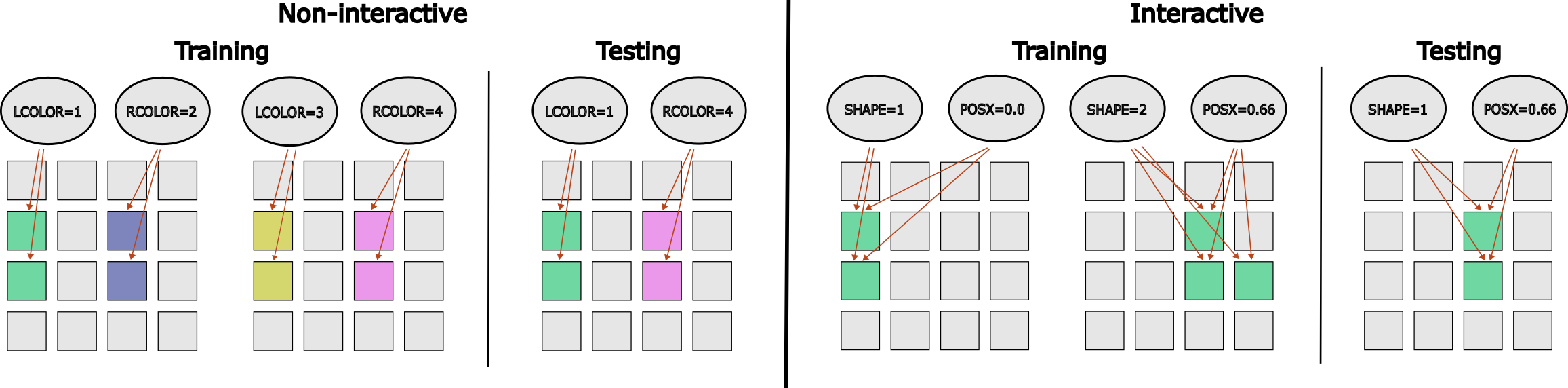}
    \caption{\textbf{Non-interactive and interactive combinations as graphs}. The panels on the left show an example of two factors \comb{lcolor, rcolor} that we call non-interactive. In this case, the value of each factor determines the colour of mutually-exclusive set of pixels. The model must learn the dependencies (edges) between the generative factors and pixels. Compare this to the three panels on the right, illustrating the interactive condition. In this case, the two factors \comb{shape, posX} jointly determine the value of each pixel. In this case, the model must learn how the value of shape nodes non-linearly determines the value of pixel nodes based on the value of posX node.}
    \label{fig:interactive}
    \vspace{-0.5cm}
\end{figure}

We illustrate these two types of combinations in Figure~\ref{fig:interactive}. In the first case ( e.g., \comb{floor-hue, wall-hue}), two factors determine mutually exclusive pixels of the image. We call these types of combinations \textit{non-interactive}. Solving the combinatorial generalisation problem in the non-interactive case is trivial -- the model can do this by learning independent mappings between sets of pixels and the corresponding generative factors. In the second case (e.g., \comb{shape, posX}), the value of pixels is jointly determined by two factors in a nonlinear and complex manner. A pixel node may have value $+1$ for a given value of \comb{shape} and \comb{posX}, but this value may change to $-1$ for a slight change in either factor. Other changes will have no effect on the value of some pixels. We call these types of combinations \textit{interactive}. To succeed at combinatorial generalisation in the interactive condition, a model must not only learn the mapping between pixels and factors, but also learn how the factors combine to jointly determine the value of each pixel.

Based on this insight, we hypothesised that models may fail to perform combinatorial generalisation in the interactive condition, but succeed in the non-interactive condition. To test this hypothesis, we carried out a set of experiments where the model had to solve the problem of combinatorial generalisation, but the combinations that were excluded did not interact in the training examples. We can do this for both the \texttt{3DShapes} dataset as well as the \texttt{MPI3D} dataset, as both datasets involve images where the canvas contains several disjoint elements. So, we repeated the semi-supervised learning experiments above, where models had to learn the composition task, but instead of excluding combinations where the generative factors interacted with each other, we excluded the following combinations from the training set:
\begin{itemize}[noitemsep,topsep=0pt,leftmargin=10pt]
    \item \texttt{\textbf{3DShapes}}: Exclude all combinations such that \comb{floor-hue~$<0.25$, wall hue~$>0.75$}: none of the training images had the combination of floors with a ``warm'' hue and walls with a ``cold'' hue.
    \item \texttt{\textbf{MPI3D}}: Exclude all combinations such that \comb{shape=\{cylinder,sphere\}, background color=salmon}. Note, even though shape combinations are excluded, the combination of generative factors excluded do not interact (i.e. determine different parts of the image).
\end{itemize}

It is also possible that model struggled with factors that combine with \comb{shape} not because it is an interactive factor, but because it is a `discrete' factor, taking on a few specific and unrelated values. If this is the case, then a model should be able to succeed at combinatorial generalisation by learning discrete latent representations. A recent model -- CascadeVAE \citep{jeong2019learning} -- addresses this by concurrently inferring continuous and discrete latent factors. We tested this model on the image reconstruction task in the recombination-to-range condition of \texttt{dSprites} dataset.

Finally, it is possible that standard VAEs struggle with the combinatorial generalisation because they cannot capture the dependencies between generative factors. A recent model -- Commutative LieGroupVAE \citep{zhu_commutative_2021} -- uses an adaptive equivariant structure, rather than a fixed vector space, to learn factors of variation in the data. This approach combines explicit modeling group operations plus penalties to the learned basis in order to learn a highly disentangled representation of the input (see the original work for more details). The hope is that because this method is adaptive, it may be able to capture not only the generative factors underlying the data, but also the dependencies between them (see Figure~\ref{fig:interactive}). As with CascadeVAE, we use the same training configuration as in the original work and test on the interactive conditions of dSprites and Shapes3D datasets (see details in Appendix~\ref{app:lie-group-vae}).

\subsection{Results}

\begin{figure*}[!h]
\centering
\includegraphics[width=0.75\textwidth]{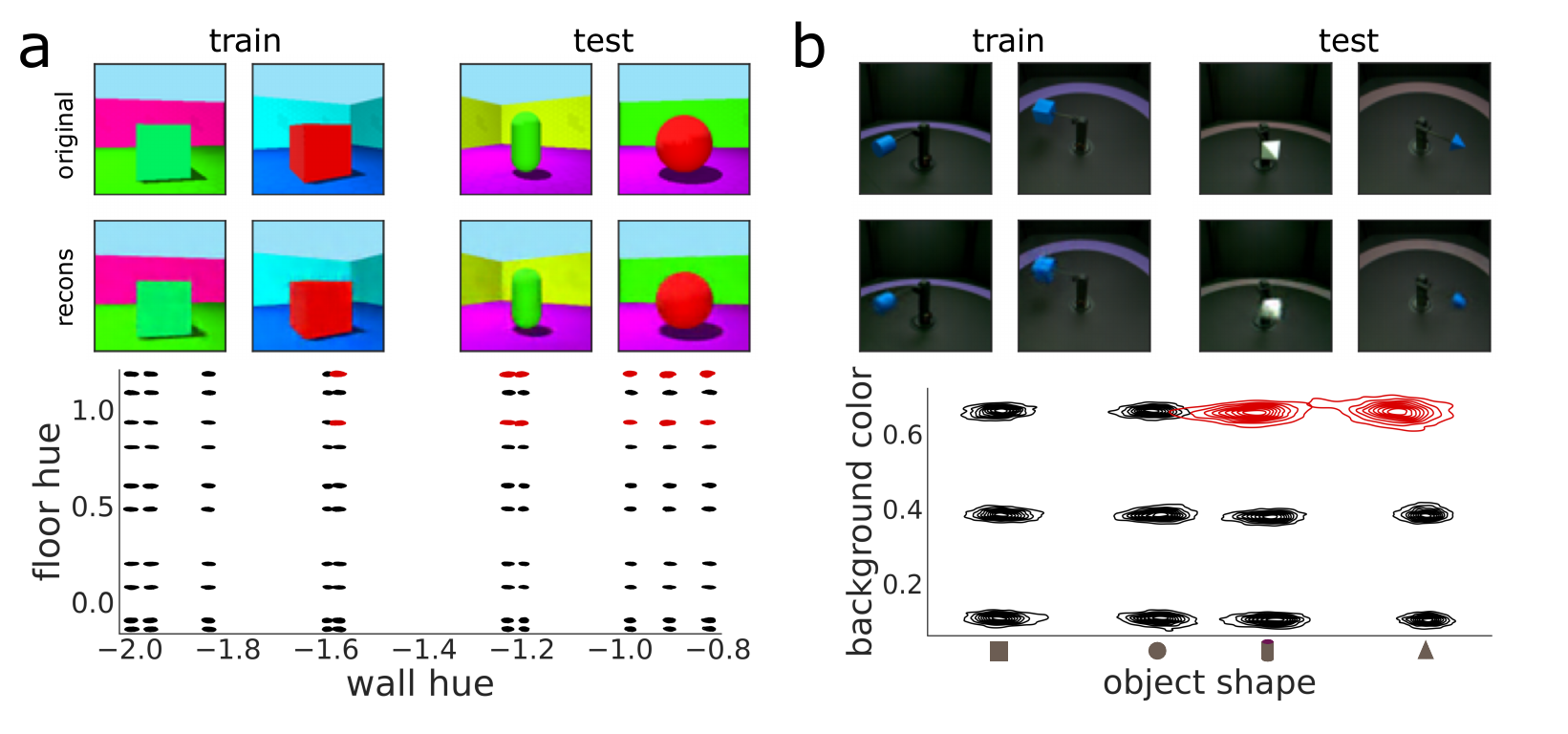}
\caption{\textbf{Latent space for conditions where models succeed} Visualisation of the latent space for models trained on conditions in which they succeed at generalisation. The figure follows the same conventions as in Figure~\ref{fig:results-comp}. (a) Results for 3DShapes when removing floor and wall color combinations. (b) Visualisation for the MPI3D dataset for combinations where the salmon background has not been seen with the cylinder shape.}
\label{fig:success-results}
\vspace{-0.8cm}
\end{figure*}

For the non-interactive conditons, we again observed that models learned highly disentangled representations for this task. More interestingly, we also observed that models now succeeded at the combinatorial generalisation task. Figure~\ref{fig:success-results} shows some typical reconstructions for training as well as test images. These reconstructions successfully reproduce the unseen combination of floor hues and wall hues for the \texttt{3DShapes} dataset and the combination of shape and background arc color for the \texttt{MPI3D} dataset. This figure also shows the joint probability distribution for trained as well as novel combination of latent values. In contrast to Figure~\ref{fig:results-comp}, we now see that the means of the test latent distributions (in red) fall in the expected location and a show a much smaller variance.

We also observed that alternative models cannot solve the hard generalisation challenges described (See Appendix~\ref{app:controls}). Figure~\ref{fig:cascade-dsprites} shows these results for CascadeVAE on dSprites (the only dataset for which authors provide training parameters), and how the model clearly fails at combinatorial generalisation as before, unable to map the previously seen shape to a novel position. Figure~\ref{fig:lie-interactive} shows the same results for LieGroupVAE, with model succeeding at the non-interactive condition in Shapes3D but failing the interactive one. 
\vspace{-0.4cm}

\section{Discussion}\label{sec:discus}
\vspace{-0.2cm}

Our world is inherently compositional -- it can be decomposed into simpler parts and relationships between these parts. The idea behind learning disentangled representations is to recover this underlying compositional structure of the world from perceptual inputs. Unsupervised learning models, such as VAEs, aim to do this by separating out the factors that remain invariant under transformations of other factors \citep{higgins_towards_2018}. It is therefore tempting to conclude that models that manage to show a high degree of disentanglement on the training set also capture the compositional structure of the world. If they did, they should be able to generalise to settings that present novel combination of the factors of variation. However, in our experiments, we observed that models that learned highly disentangled representations nevertheless failed at combinatorial generalisation, not only in the reconstruction space, but also in the latent space. These failures reproduced over several different datasets, different types of decoders, different loss functions and under a variety of different task settings.

Our interpretation of these results is that models manage to achieve a high degree of disentanglement by discovering factors that remain invariant over training examples and simply associating perceptual inputs with these factors. However, to capture the compositional structure of the world, models must additionally understand how factors interact to cause the perceptual input -- that is, develop a good causal model of the world. We think the failure of models to form such a causal model is the reason why models succeeded at problems of combinatorial generalisation that do not involve an \emph{interaction} between the left-out factors (Figure~\ref{fig:success-results}), but consistently failed at problems where these factors interacted (Figure~\ref{fig:results-comp}). When generative factors do not interact, the model does not need to learn how the factors combine to determine the same part of the output space. Instead, it can simply map the value of each generative factor to a different location. It \emph{appears} to solve combinatorial generalisation, but it does not understand how generative factors combine.

In summary, our work shows that the problem of combinatorial generalisation remains unsolved in both latent space and reconstruction space, even for highly disentangled models. By highlighting this limitation, we hope to inspire more work exploring new approaches that emulate this key capacity of human cognition, thus endowing models with a better understanding of the compositional structure of our world.

\section*{Acknowledgments}
\vspace{-0.4cm}
The authors would like to thank the members of the Mind \& Machine Learning Group for useful comments throughout the different stages of this research. This research was supported by a ERC Advanced Grant, Generalization in Mind and Machine \#741134.

\section*{References}

{\def\section *#1{\small} 

\bibliographystyle{unsrtnat}
\bibliography{main}

\begin{thebibliography}{37}
\providecommand{\natexlab}[1]{#1}
\providecommand{\url}[1]{\texttt{#1}}
\expandafter\ifx\csname urlstyle\endcsname\relax
  \providecommand{\doi}[1]{doi: #1}\else
  \providecommand{\doi}{doi: \begingroup \urlstyle{rm}\Url}\fi

\bibitem[Higgins et~al.(2017)Higgins, Matthey, Pal, Burgess, Glorot, Botvinick,
  Mohamed, and Lerchner]{higgins2017}
Irina Higgins, Loic Matthey, Arka Pal, Christopher Burgess, Xavier Glorot,
  Matthew Botvinick, Shakir Mohamed, and Alexander Lerchner.
\newblock $\beta$-{VAE}: {Learning} basic visual concepts with a constrained
  variational framework.
\newblock page~13, 2017.

\bibitem[Burgess et~al.(2018)Burgess, Higgins, Pal, Matthey, Watters,
  Desjardins, and Lerchner]{burgess_understanding_2018}
Christopher~P. Burgess, Irina Higgins, Arka Pal, Loic Matthey, Nick Watters,
  Guillaume Desjardins, and Alexander Lerchner.
\newblock Understanding disentangling in \${\textbackslash}beta\$-{VAE}.
\newblock \emph{arXiv:1804.03599 [cs, stat]}, April 2018.
\newblock URL \url{http://arxiv.org/abs/1804.03599}.
\newblock arXiv: 1804.03599.

\bibitem[Kim and Mnih(2019)]{kim_disentangling_2019}
Hyunjik Kim and Andriy Mnih.
\newblock Disentangling by {Factorising}.
\newblock \emph{arXiv:1802.05983 [cs, stat]}, July 2019.
\newblock URL \url{http://arxiv.org/abs/1802.05983}.
\newblock arXiv: 1802.05983.

\bibitem[Locatello et~al.(2020)Locatello, Poole, R{\"a}tsch, Sch{\"o}lkopf,
  Bachem, and Tschannen]{locatello2020weakly}
Francesco Locatello, Ben Poole, Gunnar R{\"a}tsch, Bernhard Sch{\"o}lkopf,
  Olivier Bachem, and Michael Tschannen.
\newblock Weakly-supervised disentanglement without compromises.
\newblock In \emph{International Conference on Machine Learning}, pages
  6348--6359. PMLR, 2020.

\bibitem[Lin et~al.(2020)Lin, Thekumparampil, Fanti, and Oh]{lin2020infogan}
Zinan Lin, Kiran Thekumparampil, Giulia Fanti, and Sewoong Oh.
\newblock Infogan-cr and modelcentrality: Self-supervised model training and
  selection for disentangling gans.
\newblock In \emph{International Conference on Machine Learning}, pages
  6127--6139. PMLR, 2020.

\bibitem[Watters et~al.(2019)Watters, Matthey, Burgess, and
  Lerchner]{watters_spatial_2019}
Nicholas Watters, Loic Matthey, Christopher~P. Burgess, and Alexander Lerchner.
\newblock Spatial {Broadcast} {Decoder}: {A} {Simple} {Architecture} for
  {Learning} {Disentangled} {Representations} in {VAEs}.
\newblock \emph{arXiv:1901.07017 [cs, stat]}, August 2019.
\newblock URL \url{http://arxiv.org/abs/1901.07017}.
\newblock arXiv: 1901.07017.

\bibitem[Klindt et~al.(2020)Klindt, Schott, Sharma, Ustyuzhaninov, Brendel,
  Bethge, and Paiton]{klindt2020towards}
David Klindt, Lukas Schott, Yash Sharma, Ivan Ustyuzhaninov, Wieland Brendel,
  Matthias Bethge, and Dylan Paiton.
\newblock Towards nonlinear disentanglement in natural data with temporal
  sparse coding.
\newblock \emph{arXiv preprint arXiv:2007.10930}, 2020.

\bibitem[Duan et~al.(2020)Duan, Matthey, Saraiva, Watters, Burgess, Lerchner,
  and Higgins]{duan_unsupervised_2020}
Sunny Duan, Loic Matthey, Andre Saraiva, Nicholas Watters, Christopher~P.
  Burgess, Alexander Lerchner, and Irina Higgins.
\newblock Unsupervised {Model} {Selection} for {Variational} {Disentangled}
  {Representation} {Learning}.
\newblock \emph{arXiv:1905.12614 [cs, stat]}, February 2020.
\newblock URL \url{http://arxiv.org/abs/1905.12614}.
\newblock arXiv: 1905.12614.

\bibitem[Higgins et~al.(2018{\natexlab{a}})Higgins, Pal, Rusu, Matthey,
  Burgess, Pritzel, Botvinick, Blundell, and Lerchner]{higgins_darla_2018}
Irina Higgins, Arka Pal, Andrei~A. Rusu, Loic Matthey, Christopher~P. Burgess,
  Alexander Pritzel, Matthew Botvinick, Charles Blundell, and Alexander
  Lerchner.
\newblock {DARLA}: {Improving} {Zero}-{Shot} {Transfer} in {Reinforcement}
  {Learning}.
\newblock \emph{arXiv:1707.08475 [cs, stat]}, June 2018{\natexlab{a}}.
\newblock URL \url{http://arxiv.org/abs/1707.08475}.
\newblock arXiv: 1707.08475.

\bibitem[Higgins et~al.(2018{\natexlab{b}})Higgins, Sonnerat, Matthey, Pal,
  Burgess, Bosnjak, Shanahan, Botvinick, Hassabis, and
  Lerchner]{higgins_scan_2018}
Irina Higgins, Nicolas Sonnerat, Loic Matthey, Arka Pal, Christopher~P.
  Burgess, Matko Bosnjak, Murray Shanahan, Matthew Botvinick, Demis Hassabis,
  and Alexander Lerchner.
\newblock {SCAN}: {Learning} {Hierarchical} {Compositional} {Visual}
  {Concepts}.
\newblock \emph{arXiv:1707.03389 [cs, stat]}, June 2018{\natexlab{b}}.
\newblock URL \url{http://arxiv.org/abs/1707.03389}.
\newblock arXiv: 1707.03389.

\bibitem[van Steenkiste et~al.(2019)van Steenkiste, Schmidhuber, Locatello, and
  Bachem]{van_steenkiste_are_nodate}
Sjoerd van Steenkiste, Jürgen Schmidhuber, Francesco Locatello, and Olivier
  Bachem.
\newblock Are {Disentangled} {Representations} {Helpful} for {Abstract}
  {Visual} {Reasoning}?
\newblock page~14, 2019.

\bibitem[Sch{\"o}lkopf et~al.(2021)Sch{\"o}lkopf, Locatello, Bauer, Ke,
  Kalchbrenner, Goyal, and Bengio]{scholkopf2021toward}
Bernhard Sch{\"o}lkopf, Francesco Locatello, Stefan Bauer, Nan~Rosemary Ke, Nal
  Kalchbrenner, Anirudh Goyal, and Yoshua Bengio.
\newblock Toward causal representation learning.
\newblock \emph{Proceedings of the IEEE}, 109\penalty0 (5):\penalty0 612--634,
  2021.

\bibitem[Von~Humboldt et~al.(1999)Von~Humboldt, von Humboldt,
  et~al.]{von1999humboldt}
Wilhelm Von~Humboldt, Wilhelm~Freiherr von Humboldt, et~al.
\newblock \emph{Humboldt:'On language': On the diversity of human language
  construction and its influence on the mental development of the human
  species}.
\newblock Cambridge University Press, 1999.

\bibitem[Chomsky(2014)]{chomsky2014aspects}
Noam Chomsky.
\newblock \emph{Aspects of the Theory of Syntax}, volume~11.
\newblock MIT press, 2014.

\bibitem[Smolensky(1988)]{smolensky_connectionism_1988}
Paul Smolensky.
\newblock \emph{Connectionism, constituency, and the language of thought}.
\newblock University of Colorado at Boulder, 1988.

\bibitem[McCoy et~al.(2021)McCoy, Culbertson, Smolensky, and
  Legendre]{mccoy2021infinite}
R~Thomas McCoy, Jennifer Culbertson, Paul Smolensky, and G{\'e}raldine
  Legendre.
\newblock Infinite use of finite means? evaluating the generalization of center
  embedding learned from an artificial grammar.
\newblock 2021.

\bibitem[Battaglia et~al.(2018)Battaglia, Hamrick, Bapst, Sanchez-Gonzalez,
  Zambaldi, Malinowski, Tacchetti, Raposo, Santoro, Faulkner,
  et~al.]{battaglia2018relational}
Peter~W Battaglia, Jessica~B Hamrick, Victor Bapst, Alvaro Sanchez-Gonzalez,
  Vinicius Zambaldi, Mateusz Malinowski, Andrea Tacchetti, David Raposo, Adam
  Santoro, Ryan Faulkner, et~al.
\newblock Relational inductive biases, deep learning, and graph networks.
\newblock \emph{arXiv preprint arXiv:1806.01261}, 2018.

\bibitem[Montero et~al.(2020)Montero, Ludwig, Costa, Malhotra, and
  Bowers]{montero2020role}
Milton~Llera Montero, Casimir~JH Ludwig, Rui~Ponte Costa, Gaurav Malhotra, and
  Jeffrey Bowers.
\newblock The role of disentanglement in generalisation.
\newblock In \emph{International Conference on Learning Representations}, 2020.

\bibitem[Schott et~al.(2021)Schott, von K{\"u}gelgen, Tr{\"a}uble, Gehler,
  Russell, Bethge, Sch{\"o}lkopf, Locatello, and Brendel]{schott2021visual}
Lukas Schott, Julius von K{\"u}gelgen, Frederik Tr{\"a}uble, Peter Gehler,
  Chris Russell, Matthias Bethge, Bernhard Sch{\"o}lkopf, Francesco Locatello,
  and Wieland Brendel.
\newblock Visual representation learning does not generalize strongly within
  the same domain.
\newblock \emph{arXiv preprint arXiv:2107.08221}, 2021.

\bibitem[Hyvarinen and Morioka(2016)]{hyvarinen2016unsupervised}
Aapo Hyvarinen and Hiroshi Morioka.
\newblock Unsupervised feature extraction by time-contrastive learning and
  nonlinear ica.
\newblock \emph{Advances in Neural Information Processing Systems},
  29:\penalty0 3765--3773, 2016.

\bibitem[Kingma and Welling(2013)]{kingma_auto-encoding_2013}
Diederik~P. Kingma and Max Welling.
\newblock Auto-{Encoding} {Variational} {Bayes}.
\newblock \emph{arXiv:1312.6114 [cs, stat]}, December 2013.
\newblock URL \url{http://arxiv.org/abs/1312.6114}.
\newblock arXiv: 1312.6114.

\bibitem[Tolstikhin et~al.(2017)Tolstikhin, Bousquet, Gelly, and
  Schoelkopf]{tolstikhin2017wasserstein}
Ilya Tolstikhin, Olivier Bousquet, Sylvain Gelly, and Bernhard Schoelkopf.
\newblock Wasserstein auto-encoders.
\newblock \emph{arXiv preprint arXiv:1711.01558}, 2017.

\bibitem[Matthey et~al.(2017)Matthey, Higgins, Hassabis, and
  Lerchner]{matthey_dsprites_2017}
Loic Matthey, Irina Higgins, Demis Hassabis, and Alexander Lerchner.
\newblock \emph{{dSprites}: {Disentanglement} testing {Sprites} dataset}.
\newblock 2017.
\newblock URL \url{https://github.com/deepmind/dsprites-dataset/}.

\bibitem[Burgess and Kim(2018)]{3dshapes18}
Chris Burgess and Hyunjik Kim.
\newblock 3d shapes dataset.
\newblock https://github.com/deepmind/3dshapes-dataset/, 2018.

\bibitem[Gondal et~al.(2019)Gondal, Wuthrich, Miladinovic, Locatello, Breidt,
  Volchkov, Akpo, Bachem, Sch\"{o}lkopf, and Bauer]{gondal2019}
Muhammad~Waleed Gondal, Manuel Wuthrich, Djordje Miladinovic, Francesco
  Locatello, Martin Breidt, Valentin Volchkov, Joel Akpo, Olivier Bachem,
  Bernhard Sch\"{o}lkopf, and Stefan Bauer.
\newblock On the transfer of inductive bias from simulation to the real world:
  a new disentanglement dataset.
\newblock In H.~Wallach, H.~Larochelle, A.~Beygelzimer, F.~d\textquotesingle
  Alch\'{e}-Buc, E.~Fox, and R.~Garnett, editors, \emph{Advances in Neural
  Information Processing Systems}, volume~32. Curran Associates, Inc., 2019.
\newblock URL
  \url{https://proceedings.neurips.cc/paper/2019/file/d97d404b6119214e4a7018391195240a-Paper.pdf}.

\bibitem[Eastwood and Williams(2018)]{eastwood_framework_2018}
Cian Eastwood and Christopher~KI Williams.
\newblock A framework for the quantitative evaluation of disentangled
  representations.
\newblock In \emph{International Conference on Learning Representations},
  page~15, 2018.

\bibitem[Jeong and Song(2019)]{jeong2019learning}
Yeonwoo Jeong and Hyun~Oh Song.
\newblock Learning discrete and continuous factors of data via alternating
  disentanglement.
\newblock In \emph{International Conference on Machine Learning}, pages
  3091--3099. PMLR, 2019.

\bibitem[Zhu et~al.(2021)Zhu, Xu, and Tao]{zhu_commutative_2021}
Xinqi Zhu, Chang Xu, and Dacheng Tao.
\newblock Commutative {Lie} {Group} {VAE} for {Disentanglement} {Learning}.
\newblock In \emph{Proceedings of the 38th {International} {Conference} on
  {Machine} {Learning}}, pages 12924--12934. PMLR, July 2021.
\newblock URL \url{https://proceedings.mlr.press/v139/zhu21f.html}.
\newblock ISSN: 2640-3498.

\bibitem[Higgins et~al.(2018{\natexlab{c}})Higgins, Amos, Pfau, Racaniere,
  Matthey, Rezende, and Lerchner]{higgins_towards_2018}
Irina Higgins, David Amos, David Pfau, Sebastien Racaniere, Loic Matthey,
  Danilo Rezende, and Alexander Lerchner.
\newblock Towards a {Definition} of {Disentangled} {Representations}.
\newblock \emph{arXiv:1812.02230 [cs, stat]}, December 2018{\natexlab{c}}.
\newblock URL \url{http://arxiv.org/abs/1812.02230}.
\newblock arXiv: 1812.02230.

\bibitem[Kingma and Ba(2017)]{kingma_adam_2017}
Diederik~P. Kingma and Jimmy Ba.
\newblock Adam: {A} {Method} for {Stochastic} {Optimization}.
\newblock \emph{arXiv:1412.6980 [cs]}, January 2017.
\newblock URL \url{http://arxiv.org/abs/1412.6980}.
\newblock arXiv: 1412.6980.

\bibitem[Munkres(1957)]{munkres1957algorithms}
James Munkres.
\newblock Algorithms for the assignment and transportation problems.
\newblock \emph{Journal of the society for industrial and applied mathematics},
  5\penalty0 (1):\penalty0 32--38, 1957.

\bibitem[Paszke et~al.(2019)Paszke, Gross, Massa, Lerer, Bradbury, Chanan,
  Killeen, Lin, Gimelshein, Antiga, Desmaison, Kopf, Yang, DeVito, Raison,
  Tejani, Chilamkurthy, Steiner, Fang, Bai, and Chintala]{pytorch2019}
Adam Paszke, Sam Gross, Francisco Massa, Adam Lerer, James Bradbury, Gregory
  Chanan, Trevor Killeen, Zeming Lin, Natalia Gimelshein, Luca Antiga, Alban
  Desmaison, Andreas Kopf, Edward Yang, Zachary DeVito, Martin Raison, Alykhan
  Tejani, Sasank Chilamkurthy, Benoit Steiner, Lu~Fang, Junjie Bai, and Soumith
  Chintala.
\newblock Pytorch: An imperative style, high-performance deep learning library.
\newblock In H.~Wallach, H.~Larochelle, A.~Beygelzimer, F.~d\textquotesingle
  Alch\'{e}-Buc, E.~Fox, and R.~Garnett, editors, \emph{Advances in Neural
  Information Processing Systems 32}, pages 8024--8035. Curran Associates,
  Inc., 2019.

\bibitem[Harris et~al.(2020)Harris, Millman, van~der Walt, Gommers, Virtanen,
  Cournapeau, Wieser, Taylor, Berg, Smith, Kern, Picus, Hoyer, van Kerkwijk,
  Brett, Haldane, del R{\'{i}}o, Wiebe, Peterson, G{\'{e}}rard-Marchant,
  Sheppard, Reddy, Weckesser, Abbasi, Gohlke, and Oliphant]{harris2020array}
Charles~R. Harris, K.~Jarrod Millman, St{\'{e}}fan~J. van~der Walt, Ralf
  Gommers, Pauli Virtanen, David Cournapeau, Eric Wieser, Julian Taylor,
  Sebastian Berg, Nathaniel~J. Smith, Robert Kern, Matti Picus, Stephan Hoyer,
  Marten~H. van Kerkwijk, Matthew Brett, Allan Haldane, Jaime~Fern{\'{a}}ndez
  del R{\'{i}}o, Mark Wiebe, Pearu Peterson, Pierre G{\'{e}}rard-Marchant,
  Kevin Sheppard, Tyler Reddy, Warren Weckesser, Hameer Abbasi, Christoph
  Gohlke, and Travis~E. Oliphant.
\newblock Array programming with {NumPy}.
\newblock \emph{Nature}, 585\penalty0 (7825):\penalty0 357--362, September
  2020.
\newblock \doi{10.1038/s41586-020-2649-2}.
\newblock URL \url{https://doi.org/10.1038/s41586-020-2649-2}.

\bibitem[Fomin et~al.(2020)Fomin, Anmol, Desroziers, J., and
  Tejani]{pytorch-ignite}
V.~Fomin, J.~Anmol, S.~Desroziers, Kriss J., and A.~Tejani.
\newblock High-level library to help with training neural networks in pytorch.
\newblock \url{https://github.com/pytorch/ignite}, 2020.

\bibitem[{K}laus {G}reff et~al.(2017){K}laus {G}reff, {A}aron {K}lein, {M}artin
  {C}hovanec, {F}rank {H}utter, and {J}\"urgen
  {S}chmidhuber]{klaus_greff-proc-scipy-2017}
{K}laus {G}reff, {A}aron {K}lein, {M}artin {C}hovanec, {F}rank {H}utter, and
  {J}\"urgen {S}chmidhuber.
\newblock {T}he {S}acred {I}nfrastructure for {C}omputational {R}esearch.
\newblock In {K}aty {H}uff, {D}avid {L}ippa, {D}illon {N}iederhut, and {M}
  {P}acer, editors, \emph{{P}roceedings of the 16th {P}ython in {S}cience
  {C}onference}, pages 49 -- 56, 2017.
\newblock \doi{10.25080/shinma-7f4c6e7-008}.

\bibitem[Hunter(2007)]{Hunter:2007}
J.~D. Hunter.
\newblock Matplotlib: A 2d graphics environment.
\newblock \emph{Computing in Science \& Engineering}, 9\penalty0 (3):\penalty0
  90--95, 2007.
\newblock \doi{10.1109/MCSE.2007.55}.

\bibitem[Waskom(2021)]{Waskom2021}
Michael~L. Waskom.
\newblock seaborn: statistical data visualization.
\newblock \emph{Journal of Open Source Software}, 6\penalty0 (60):\penalty0
  3021, 2021.
\newblock \doi{10.21105/joss.03021}.
\newblock URL \url{https://doi.org/10.21105/joss.03021}.

\end{thebibliography}

}
\newpage

\newpage


\section*{Supplementary Material}

\appendix

\section{Experimental setup}
\subsection{Composition task}\label{app:task}

Here we describe the image composition task in more detail. This task was created by \citet{montero2020role} to check whether a different task than simple image reconstruction helps the model to learn more disentangled representations and as a consequence show combinatorial generalisation. At a high level the task is a manipulation task, where models must learn to extract the relevant factors $g_{i}$ from two images, $X_{og}$ and $X_{trans}$, and combine them based on a given `action':
\begin{align*}
    \textbf{input} &= X_{og}, \; X_{trans}, \; \mathbf{q} \\
    \textbf{output} &= X_{out}
\end{align*}
where $X_{out}$ is the output image and $\mathbf{q}$ is the query vector, that encodes the action. Following \citeauthor{montero2020role}, we use actions that involve replacing one of the properties (generative factors) of $X_{og}$ with a property of $X_{trans}$. The query vector uses a one-hot encoding of the generative factor that must be changed. See Figure~\ref{fig:task-model-2} for an example of this task.

A critical aspect of this task is that it requires the change to be only one generative factor for each trial. Restricting the manipulation to a single generative factor encourages sparseness of representations and provides a strong inductive bias for the model to learn disentangled representations. Previous research has shown that sparse transitions following natural image statistics provide a strong inductive bias towards disentanglement (see, for example, \citet{klindt2020towards}). This is because breaking the i.i.d. assumption, present in most datasets, allows for the identification of the underlying factors (see \citet{klindt2020towards} for details and \citet{hyvarinen2016unsupervised} for a similar argument for selecting independent components from data).

A second inductive bias towards disentanglement is provided by the nature of the composition task. In contrast to a reconstruction task, that doesn't provide any supervision signal, and a classification task, which provides a strong supervision signal, the composition task uses a query vector to provide a \emph{weak supervision} signal. Two studies have shown that a weak supervision signal can provide a strong inductive bias for generalisation. \citet{locatello2020weakly} showed that a few labeled examples were enough to significantly improve the disentanglement in generative models. In the second study, \citet{lin2020infogan} developed a task with a weak supervision signal, where the target was to identify the factor that had changed between two images. They used this task to train a GAN and show it leads to more disentangled representations. The query vector in the composition task constitutes a similar weak supervision signal, but provided as an input instead of as a target.

\subsubsection{Procedure}
When training models, we sampled each combination in an online fashion. The procedure works as follows:

\begin{enumerate}
    \item Sample an image $X_{og}$ and with generative factors given by the vector $\mathbf{g_{og}}$ from the dataset.
    \item Sample an action, $a$, that indexes the set of all generative factors, $\{g_{1}, \dots, g_{n}\}$.
    \item Sample a second image $X_{trans}$ such that $\mathbf{g_{trans}}[a]$, the $a$th generative factor of $X_{trans}$ does not match $\mathbf{g_{og}}[a]$, the $a$th generative factor of $X_{og}$.
    \item Compute $\mathbf{g_{out}}$ by replacing $\mathbf{g_{og}}[a]$ with $\mathbf{g_{trans}}[a]$ and get and it's associated image $X_{out}$.
\end{enumerate}

All sampling is done uniformly and we allow sampling of categorical variables such as shape.

\subsection{Models}\label{app:models}
\subsubsection{Composition operation}
We use the general architecture described in Figure~\ref{fig:task-model-2}(b) to define a model that solves the composition task. As mentioned in the main text this requires the definition of the composition operation, $f(\cdot)$ in latent space. In general, such operation can be defined as:
\begin{equation*}
    \mathbf{z_{out}} \triangleq f(\mathbf{z_{og}}, \mathbf{z_{trans}}, \mathbf{q})
\end{equation*}
where $\mathbf{z_{out}}$, $\mathbf{z_{og}}$ and $\mathbf{z_{trans}}$ are the latent representations corresponding to the output, original, and transformation images. and $\mathbf{q}$ is the one-hot vector encoding the action to be performed. 

We defined 3 different versions of this operation:
\begin{align}
    f_{mlp}(\mathbf{z_{og}}, \mathbf{z_{trans}}, \mathbf{q}) &= \mathbf{W}_{out} \cdot \text{ReLU}(\mathbf{W}[\mathbf{z_{og}};\mathbf{z_{trans}};\mathbf{q}] + \mathbf{b}) \\
    f_{lin}(\mathbf{z_{og}},\mathbf{z_{trans}}, \mathbf{q}) &= W_{out} \cdot [\mathbf{z_{og}};\mathbf{z_{trans}};\mathbf{q}] \\
    f_{interp}(\mathbf{z_{og}}, \mathbf{z_{trans}}, \mathbf{q}) &= \mathbf{z_{og}} \odot (1 - \mathbf{c}) +  \mathbf{z_{trans}} \odot \mathbf{c} \label{eq:finterp}
\end{align}
where $[.;.]$ represents the concatenation of vectors, $\mathbf{c}$ is the interpolation vector of coefficients. We define $\mathbf{c}$ in two ways for equation~\ref{eq:finterp} as described in the main text:
\begin{align}
    \mathbf{c_{learn}} &= \sigma(\mathbf{W} \cdot [\mathbf{z_{og}};\mathbf{z_{trans}};\mathbf{q}] + \mathbf{b}) \label{eq:clearn} \\
    \mathbf{c_{action}} &= \text{pad}(\mathbf{q}, \text{length}(\mathbf{z}) - \text{length}(\mathbf{q})) \label{eq:caction}
\end{align}
where the first option has learnable parameters $W$ and $\mathbf{b}$. The second just pads the query vector, $\mathbf{q}$, with zeros to match the vector length of the latent representation.

Preliminary tests showed that using (1) or (2) as composition operations lead to poor disentanglement. Thus we only tested the two variants of (3) in the rest of the experiments. For some datasets such as dSprites, (5) worked better. For 3DShapes on the other hand, (4) gave better results.

For the rest of the architectures (encoders and decoders) we use the same parameters across datasets for each condition. For dSprites and 3DShapes, we use a similar architecture as in \citet{kim_disentangling_2019} but increase the number of channels in the early convolutions. MPI3D required us to increase these values again in order to achieve good reconstructions. For Circles and Simple we use the architecture proposed in \citet{watters_spatial_2019} but slightly increase the channel size in the decoder to avoid a performance overhead related to PyTorch's implementation. See Table~\ref{tab:model-params} for details.

\begin{table}[h]
    \caption{\textbf{Architecture parameters}}
    \centering
    \begin{threeparttable}
    \begin{tabular*}{0.9\textwidth}{c@{\extracolsep{\fill}}c|cc|cc}
    \toprule
    \multicolumn{2}{c|}{dSprites \& 3DShapes} &  \multicolumn{2}{c|}{MPI3D} & \multicolumn{2}{c}{Circles \& Simple}\\
    Encoder & Decoder & Encoder & Decoder & Encoder & Decoder \\
    \midrule
    Conv(32, 4, 2)  & Transpose  & Conv(64, 4, 2)  & Transpose  & Conv(64, 4, 2) & SBD(64, 64)    \\
    Conv(32, 4, 2)  & of encoder & Conv(64, 4, 2)  & of encoder & Conv(64, 4, 2) & Conv(64, 5, 1) \\
    Conv(64, 4, 2)  &            & Conv(128, 4, 2) &            & Conv(64, 4, 2) & Conv(64, 5, 1) \\
    Conv(64, 4, 2)  &            & Conv(128, 4, 2) &            & Conv(64, 4, 2) & Conv(64, 5, 1) \\
    Conv(128, 4, 2) &            & Conv(256, 4, 2) &            & Linear(256)    & Conv(64, 5, 1) \\
    Linear(256)     &            & Linear(256)     &            & Linear(20)     &                \\
    Linear(20)      &            & Linear(20)      &            &                &                \\
    \bottomrule
    \end{tabular*}
    \end{threeparttable}
    \label{tab:model-params}
\end{table}

All layers where followed by ReLU activation functions except for the last one. The latent layers used were 10-dimensional, parameterized diagonal Gaussians in all cases. For both the Conv and DeConv layer the parameters indicate number of channels, size of convolution filter and the stride. ``Same'' padding was used throughout. For completion, we note that the architectures for the latent prediction task were the same ones as the corresponding encoder.

\subsection{Datasets}

Throughout the article we use standard datasets used to test disentangled models in the literature. These are: dSprites, 3DShapes, MPI3D and Circels. Below we describe each of the generative factors they contain. We've focused on datasets that have been specifically designed for this purpose and thus contain explicit values of the generative factors. This is necessary in order to compute the level of disentanglement with most metrics, including the DCI metric.

\begin{itemize}
    \item \texttt{\textbf{dSprites}}: Introduced in Higgins, \citet{higgins2017} to test the original $\beta$-VAE approach to disentanglement. It contains the following generative factors: \comb{shape, scale, orientation, position X, position Y}. Orientation here refers to the rotation of the shape along it's center of mass. This is as opposed to the meaning in \texttt{3DShapes} (see below). The GitHub repository for \texttt{dSprites} can be found at: \url{https://github.com/deepmind/dsprites-dataset}.
    \item \texttt{\textbf{3DShapes}}: Introduced in Kim \citet{kim_disentangling_2019}, to study the FactorVAE penalty. It contains the 6 generative factors: \comb{floor hue, wall hue, object hue, scale, shape, orientation}. Colors are defined in the HSV format, and the values correspond to the hue component. Here orientation defines the angle of point-of-view for the scene. The objects themselves do not rotate. This dataset can be found at:  \url{https://github.com/deepmind/3d-shapes}.
    \item \texttt{\textbf{MPI3D}}: This dataset was proposed in Gondal et al, \citet{gondal2019} as part of the the NEURIPS Disentanglement Challenge. It contains seven generative factors: \comb{object color, object shape, object size, camera height, background color, horizontal axis, vertical axis}. We note that vertical axis and horizontal axis have complex non-linear dependencies between them in the rendered image. The names can also be misleading as horizontal axis controls the height of the arm while vertical axis controls the rotation in the direction perpendicular to the horizontal axis. The GitHub repository for this dataset can be found at: \url{https://github.com/rr-learning/disentanglement_dataset}.
    \item \texttt{\textbf{Circles}}: Introduced to test the capabilities of the Spatial Broadcast Decoder models in Watters et al, \citet{watters_spatial_2019}. It contains only two factors: \comb{position x, position y}, the model being required to only render circles at the relevant position. There is no published dataset in this case, though the authors mention that they generated this dataset using the \href{https://github.com/deepmind/spriteworld}{Spriteworld} library, which is a library designed to generate simple datasets for reinforcement learning. We adapted the library to generate still images similar to those found in \citet{watters_spatial_2019}. 
    \item \texttt{\textbf{Simple}}: We modified the Circles dataset to create the Simple Sprites dataset in order to test the effect of introducing an additional shape when testing the combinatorial generalisation capabilities of the Spatial Broadcast decoder. In addition to the two position factors in the Circles dataset, we added a shape factor that could take one of two values \comb{shape=\{triangle,square\}}. We chose these shapes so that they are sufficiently different from each other. To generate the dataset we use the same library that we use to replicate the Circles dataset above. 

\end{itemize}

\subsubsection{Combinatorial generalisation test conditions}

For each dataset we tested one success condition and one failure condition. These all constitute recombination-to-range conditions as definde in \citet{montero2020role} but we refer to them as combinatorial generalisation conditions \footnote{This is the terminology used \citet{schott2021visual}, though, technically speaking, their interpolation condition was also a combinatorial generalisation case analogous to the recombination-to-element condition in \citeauthor{montero2020role}}. The failure conditions were defined as follows

\begin{itemize}
    \item \texttt{\textbf{dSprites}}: All images such that \comb{shape=square, posX$>0.5$} were excluded from the training set. Squares never appear on the right side of the image, but do appear on the left and other shapes (hearts, ellipsis) are observed on the right as well.
    \item \texttt{\textbf{3DShapes}}: All images such that \comb{shape=pill, object-hue=$>0.5$} were excluded from the training set. Thus, pills colored as any of the colors in the second half of the HSV spectrum did not appear in the training set. These colors (shades of blue, purple, etc) were observed on the other shapes, and the pill was observed with other colors such as red and orange. Additionally we removed every other color from the dataset. This helped us clearly discretize the generative factors and latent representations, allowing us to clearly observe the performance of the models in the latent space.
    \item \texttt{\textbf{MPI3D}}: All images such that \comb{shape=cylinder, vertical axis$>0.5$} were excluded from the training set. We also excluded all images where \comb{shape=cone} or \comb{shape=hexagonal} as these shapes are very similar to the pyramid and cylinder, respectively and make it hard to access reconstruction accuracy. Furthermore, only images with \comb{horizontal axis=0} were included as rotation of both horizontal and vertical axis make it hard to find completely unseen combinations due to how much rotation of the objects is involved.
    \item \texttt{\textbf{Circles}}: All images such that \comb{posX~$0.5$,posY$>0.5$} were excluded from the training set. Models have not seen circles in the top right corner.
    \item \texttt{\textbf{Simple}}: All images such that \comb{$0.35<$~posX~$0.65$,$0.35$~posY$>0.65$,shape=triangle} were excluded from the training set. Similar to the previous condition, but now the unseen triangles lie in a patch in the middle of the screen. Squares were presented on all positions, similar to dSprites.
\end{itemize}

And the success ones as:

\begin{itemize}
    \item \texttt{\textbf{3DShapes}}: These test images presented a novel combination of generative factors, here \comb{floor hue~$<0.25$, wall hue~$>0.75$} -- that is, the model has seen all wall hues and floor hues in the range $[0, 1]$, but it has never seen a combination a floor with a hue~$<0.25$ with a wall of a hue~$>0.75$. In this case no discretization was necessary, as the task is easy enough that the model obtained good disentanglement and the visualisations were easy to understand.
    \item \texttt{\textbf{MPI3D}}: All images such that \comb{shape=\{cylinder,sphere\}, background color=salmon} were excluded from the training set. We also excluded all images where \comb{shape=cone} or \comb{shape=hexagonal} as these shapes are very similar to the pyramid and cylinder, respectively and make it hard to access reconstruction accuracy. Furthermore, only images with \comb{horizontal axis=0} were included as rotation of both horizontal violate our independence assumption. Models can appear over the background strip very often.
    \item \texttt{\textbf{Circles}}: All images such that \comb{$0.35<$~posX~$0.65$,$0.35$~posY$>0.65$} were excluded from the training set. Models have not seen circles in a patch located in the middle of the image.
\end{itemize}

\subsection{Training}\label{sec:train_obj}
To penalise the models we use the standard VAE objective \citep{kingma_auto-encoding_2013} and the Wasserstein loss (WAE, \citet{tolstikhin2017wasserstein}). In the case of the MPI3D dataset, WAEs were the only one that could successfully converge for the semi-supervised setting, which we use as a sanity check. This is likely because the VAE objective forces the distribution of the latent representation for each input to match the prior. On the other hand WAEs only require the marginal posterior \emph{over all inputs} to match the prior. The result is a latent representation that is more flexible, which seems to be required in this more complex dataset. As a consequence, we do not test VAE models on the MPI3D dataset for the composition task.

We trained the models to convergence, which in all cases occurred within 100 epochs. Batch sizes were kept at 64 and learning rates at $1e-4$ as in previous work for dSprites, 3DShapes and MPI3D. For the Circles dataset we used a batch size of 16 and a learning rate of $3e-4$ as in \citeauthor{watters_spatial_2019}. We used the Adam optimizer with the default PyTorch values for all remaining parameters \citep{kingma_adam_2017}.

\subsection{Measuring and visualizing disentanglement}\label{app:disent-def}

We measured disentanglement using the DCI metric of \citet{eastwood_framework_2018}. DCI defines three different metrics: disentanglement, completeness and informativeness. For our purposes, we are only interested in the first, but they are all computed in a similar way. Disentanglement here is defined as the degree to which a latent variable represents at most one and only one generative factor.

Let $v_j$ be the ground truth values for each factor $g_j$. Assuming a trained model, the metric works as follows:

\begin{enumerate}
    \item Compute the latent representation $z_n$ = $\text{encode}(x_n)$ for all images $x_n$ in the dataset.
    \item For each factor $g_j$ solve a regression problem with the $z_n$'s as covariates.
    \item Construct a matrix $C$, where each index, $c_{ij}$, is the absolute value of the regression coefficient between the latent variable $i$ and generative factor $j$.
    \item $C$ is then used to quantify the deviation of the latent representation from the ideal one.
\end{enumerate}

The authors propose using Lasso regression or Random Forests (they provide a method to determine the coefficient for the latter). In this work we use Lasso with their proposed hyper-parameters. To compute the disentanglement score, the coefficients in $C$ are treated as a distributions, one for each column (latent), from which the entropy is computed:

\begin{equation*}
    H(P_i) = -\sum^J_{j=0} P_{ij} \log_{J}P_{ij}
\end{equation*}
where $P_{ij} = C_{ij} / \sum_{k=0}^{J} C_{ik}$, where $J$ denotes the number of rows in $C$. The score for one variable is computed as:
\begin{equation*}
    d_i = 1 - H(P_i)
\end{equation*}
The $d_i$'s are then be averaged to obtain the final, overall disentanglement score:
\begin{align*}
    D &= \sum_{i} \rho_i \cdot d_i \\
    \rho_i &=  \frac{\sum_{j=0}^J C_{i,j}}{\sum_{l, j}C_{l, j}}
\end{align*}
where the $\rho_i$ are computed so as to account for dead units in the representation. We refer to $D$ as the disentanglement score throughout the main text, in line with most research on the topic.

\begin{figure}
    \centering
    \includegraphics[width=0.5\textwidth]{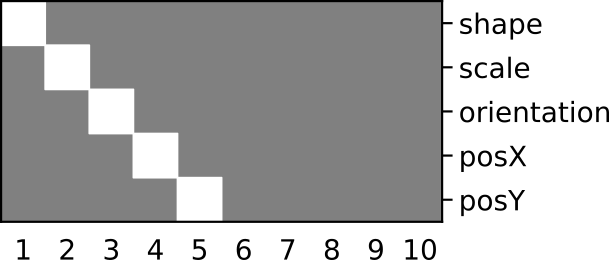}
    \caption{\textbf{Ideal disentanglement} Example of ideal disentanglement computed with the DCI metric for the dSprites dataset.}
    \label{fig:ideal-disent}
\end{figure}

DCI is interesting for a couple of reasons. First, it computes it's scores based on the ability to predict values from the latent representations. Additionally, the coefficients $C$ can be visualized using Hinton matrices, which means we can use it for both quantitative measurements and for visual inspection of the trained models, as we see below. An ideal model in this framework will produce a matrix $C$ that is diagonal, up to some permutation of the latent variables and disregarding any extra dimensions used in the latent representation (see Figure~\ref{fig:ideal-disent} for an example). This fits well with intuitions regarding disentanglement and with the definition proposed in \citet{higgins_towards_2018}. Given these coefficients, it is easy to assign latents to the corresponding generative factor. To avoid assigning a latent to two factors, we used the Munkres algorithm \citep{munkres1957algorithms} as in \citet{hyvarinen2016unsupervised}.

For a highly disentangled model we can use this assignment to visualize pairs of latent variables in a 2D plot. Given to factors of interest (say `shape' and `position') we the latent variables that were assigned to them by the above procedure for each point in the dataset. This gives us a matrix of size $Nx2$ we can then compute a kernel-density estimate of the distribution of these points. Highly disentangled models should look like a regular lattice of almost Gaussian distributions. Models that are not as disentangled can have different properties, such as drifting or increased variance of the estimated kernel densities.

\subsection{Implementation}

All the models and tasks where implemented in PyTorch (version $\geq$ 3.8\footnote{Note that 3.8 is a hard requirement here if we want to access the \texttt{\textbf{tile}} function for the Spatial Broadcast Decoder. Otherwise it must be implemented in Numpy and converted to a PyTorch tensor manually.}, \citet{pytorch2019}) and Numpy \citep{harris2020array}. The experiments where performed with the aid of the Ignite and Sacred frameworks \citep{pytorch-ignite,klaus_greff-proc-scipy-2017}. Visualizations were created using Matplotlib \citep{Hunter:2007} and Seaborn \citep{Waskom2021}. Code required to reproduce the results can be found at \url{https://github.com/mmrl/lost-in-latent-space}.

We trained a total of 48 + 9 models accounting for all combinations of datasets (4, dSprites, 3DShapes, MPI3D and Circles), losses (2, VAE and WAE), conditions (3, baseline with all the data, one success and one failure) and decoder architectures (standard or Spatial Broadcast) for the semi-supervised case, and the same combination of datasets and conditions for the latent prediction one (excluding the circles dataset). Each model took between 2 and 5 hours to train depending on the condition, except the latent-prediction case and the the circles dataset (under 1 hour). This totals to roughly 150 hours of wall clock time for training. Both this and the analysis were performed on a single workstation using an Intel Core i7-9700K CPU, with 32 GiB of RAM and an NVIDIA RTX 2080Ti GPU.

\section{Results}

Below we provide Model scores (that were used to measure success at reconstruction as well as degree of disentanglement), some example reconstructions for each type of model and Hinton matrices that are useful to visualising the degree of disentanglement. We also provide visualisations of the latent space (akin to the ones presented in the main text). In each case, we present examples of two types of training objective (VAE \& WAE -- see Section~\ref{sec:train_obj}) and two types of interpolation (fixed and learned -- see Equations~\ref{eq:caction} and~\ref{eq:clearn}) -- making a total of four models for each condition (except for the \texttt{MPI3D} dataset where only models trained with the WAE objective succeeded in learning the task). 

\subsection{Model scores}\label{app:scores}

\begin{table}[h!]
    \centering
    \caption{\textbf{Scores for dSprites failure condition} The scores for the condition \comb{shape=square,posX>0.5}. Reconstruction loss for VAEs are computed with pixel-wise binary cross-entropy (Bernoulli loss). For the WAE it is the pixel-wise mean squared error.}
    \begin{tabular}{lrrr}
    \toprule
    {} &     Train loss &        Test loss &  Disentanglement \\
    Model                &           &             &                  \\
    \midrule
    VAE + learned interp &  8.353618 &  183.369238 &         0.881088 \\
    VAE + fixed interp   &  7.524311 &  366.997949 &         \textbf{0.999999} \\
    WAE + learned interp &  2.310111 &   44.854529 &         0.917322 \\
    WAE + fixed interp   &  2.038967 &   45.108577 &         \textbf{0.983566} \\
    \bottomrule
    \end{tabular}
    \label{tab:comp-dsprites-sqr2px}
\end{table}

\begin{table}[h!]
    \centering
    \caption{\textbf{Scores for 3DShapes failure condition} The scores for the condition \comb{shape=pill,object hue=\{blue, purple\}}. Reconstruction loss for VAEs are computed with pixel-wise binary cross-entropy (Bernoulli loss). For the WAE it is the pixel-wise mean squared error.}
    \begin{tabular}{lrrr}
    \toprule
    {} &        Train loss &         Test loss &  Disentanglement \\
    Model                &              &              &                  \\
    \midrule
    VAE + learned interp &  3459.323259 &  5050.596667 &         \textbf{0.980030} \\
    VAE + fixed interp   &  3460.841481 &  5517.578000 &         \textbf{0.960200} \\
    WAE + learned interp &     5.905513 &   214.398937 &         \textbf{0.999999} \\
    WAE + fixed interp   &     7.470970 &   224.247479 &         \textbf{0.973512} \\
    \bottomrule
    \end{tabular}
    \label{tab:comp-3dshapes-shape2ohue-even-hues}
\end{table}

\begin{table}[h!]
    \centering
    \caption{\textbf{Scores for 3DShapes success condition} The scores for the condition \comb{wall hue=\{red, orange\},floor hue\{purple, magenta\}}. Reconstruction loss for VAEs are computed with pixel-wise binary cross-entropy (Bernoulli loss). For the WAE it is the pixel-wise mean squared error.}
    \begin{tabular}{lrrr}
    \toprule
    {} &        Train loss &         Test loss &  Disentanglement \\
    Model                &              &              &                  \\
    \midrule
    VAE + learned interp &  3467.451818 &  3498.955556 &         \textbf{0.975767} \\
    VAE + fixed interp   &  3471.045758 &  3510.643056 &         0.893841 \\
    WAE + learned interp &     5.079037 &     6.359890 &         \textbf{0.972172} \\
    WAE + fixed interp   &     5.932047 &    32.906424 &         0.948327 \\
    \bottomrule
    \end{tabular}
    \label{tab:comp-3dshapes-whue2fhue}
\end{table}

\begin{table}[h!]
    \centering
    \caption{\textbf{Scores for MPI3D failure condition} The scores for the condition \comb{object shape=cylinder,vertical axis >0.5}. Reconstruction loss for VAEs are computed with pixel-wise binary cross-entropy (Bernoulli loss). For the WAE it is the pixel-wise mean squared error.}
    \begin{tabular}{lrrr}
    \toprule
    {} &     Train loss &      Test loss &  Disentanglement \\
    Model                &           &           &                  \\
    \midrule
    WAE + learned interp &  2.114849 &  4.598420 &         0.675261 \\
    WAE + fixed interp   &  1.586387 &  5.227437 &         \textbf{0.976362} \\
    \bottomrule
    \end{tabular}
    \label{tab:comp-mpi3d-shape2vx}
\end{table}

\begin{table}[h!]
    \centering
    \caption{\textbf{Scores for MPI3D failure condition} The scores for the condition \comb{shape=\{cylinder,sphere\}, background color=salmon}. Reconstruction loss for VAEs are computed with pixel-wise binary cross-entropy (Bernoulli loss). For the WAE it is the pixel-wise mean squared error.}
    \begin{tabular}{lrrr}
    \toprule
    {} &     Train loss &      Test loss &  Disentanglement \\
    Model                &           &           &                  \\
    \midrule
    WAE + learned interp &  2.014464 &  2.271657 &         0.702668 \\
    WAE + fixed interp   &  1.694517 &  2.517065 &         0.966975 \\
    \bottomrule
    \end{tabular}
    \label{tab:comp-mpi3d-cyl2bkg}
\end{table}

\begin{table}[h!]
    \centering
    \caption{\textbf{Scores for the Circles dataset} The scores for the two conditions in the score dataset. The ``corner'' condition excludes instances with \comb{posX>0.5, posY>0.75}. The condition ``midpos'' excludes those with \comb{0.35<posX<0.65,0.35<posY<0.65}.}
    \begin{tabular}{llrrr}
    \toprule
    {} &       Condition &       Train loss &         Test loss &  Disentanglement \\
    Model  &             &             &              &                  \\
    \midrule
    VAE + fixed interp & corner  &  113.977726 &  4313.172778 &         \textbf{0.843828} \\
    VAE + fixed interp & midpos &  115.894145 &   157.793549 &         0.781331 \\
    \bottomrule
    \end{tabular}
    \label{tab:comp-circles}
\end{table}

\begin{table}[h!]
    \centering
    \caption{\textbf{Scores for the Simple dataset} The scores for the two conditions in the score dataset. The ``corner'' condition excludes instances with \comb{posX>0.5, posY>0.75,shape=triangle}. The condition ``midpos'' excludes those with \comb{0.35<posX<0.65,0.35<posY<0.65,shape=triangle}.}
    \begin{tabular}{llrrr}
    \toprule
    {} &       Condition &       Train loss &         Test loss &  Disentanglement \\
    Model  &             &             &              &                  \\
    \midrule
    VAE + fixed interp & corner  &  129.493923 &  523.572257 &         0.901404 \\
    VAE + fixed interp & midpos & 129.802943 &  267.656094 &         \textbf{0.936586} \\
    \bottomrule
    \end{tabular}
    \label{tab:comp-simple}
\end{table}

\FloatBarrier

\subsection{Reconstructions}\label{app:recons}%
\begin{figure}[h!]
    \centering
    \includegraphics[width=\textwidth]{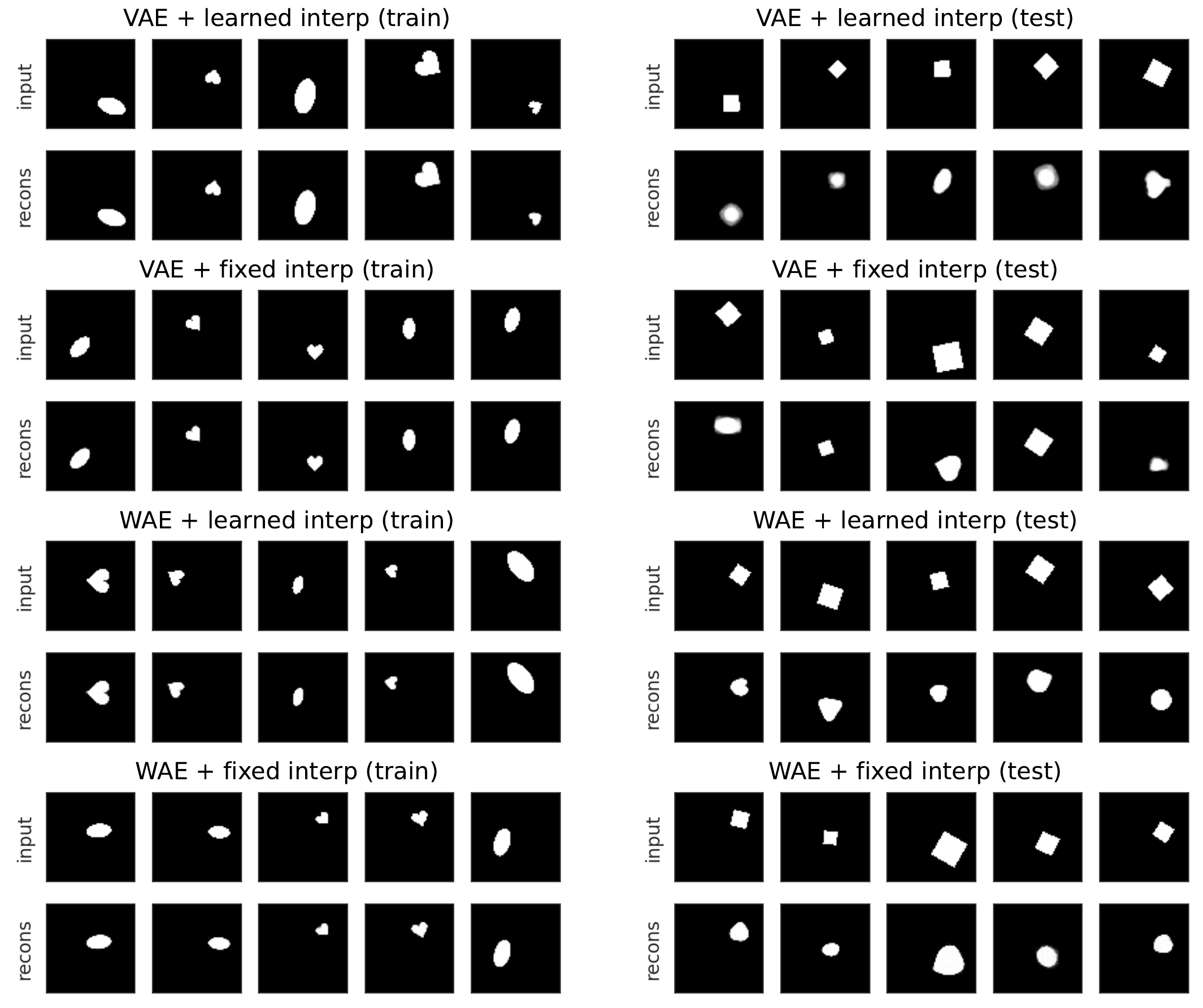}
    \caption{\textbf{Failures of combinatorial generalisation on \texttt{dSprites}} Some example reconstructions for the \texttt{dSprites} dataset for four models: (i) a VAE using learned interpolation (see Equations~\ref{eq:clearn}), (ii) a VAE using a fixed interpolation (see Equation~\ref{eq:caction}), (iii) a WAE using a learned interpolation, and (iv) a WAE using a fixed interpolation. In each case, the five figures on the left show input (first row) and reconstruction (second row) for five training images. The five figures on the right show input and reconstructions for test images. These test images presented a novel combination of generative factors, here \comb{shape=square, posX~$>0.5$} -- that is, the model has seen all shapes on both the left and right hand side of the canvas, except squares, which have only been seen on the left hand side (posX~$<0.5$). All models learn representations that are highly disentangled. While reconstructions on the training dataset are really accurate - it is clear to see that the models struggle to reconstruct the test images, where a square is presented on the right hand side of the canvas.}
    \label{fig:sqr2px-recons}
\end{figure}

\begin{figure}[h!]
    \centering
    \includegraphics[width=\textwidth]{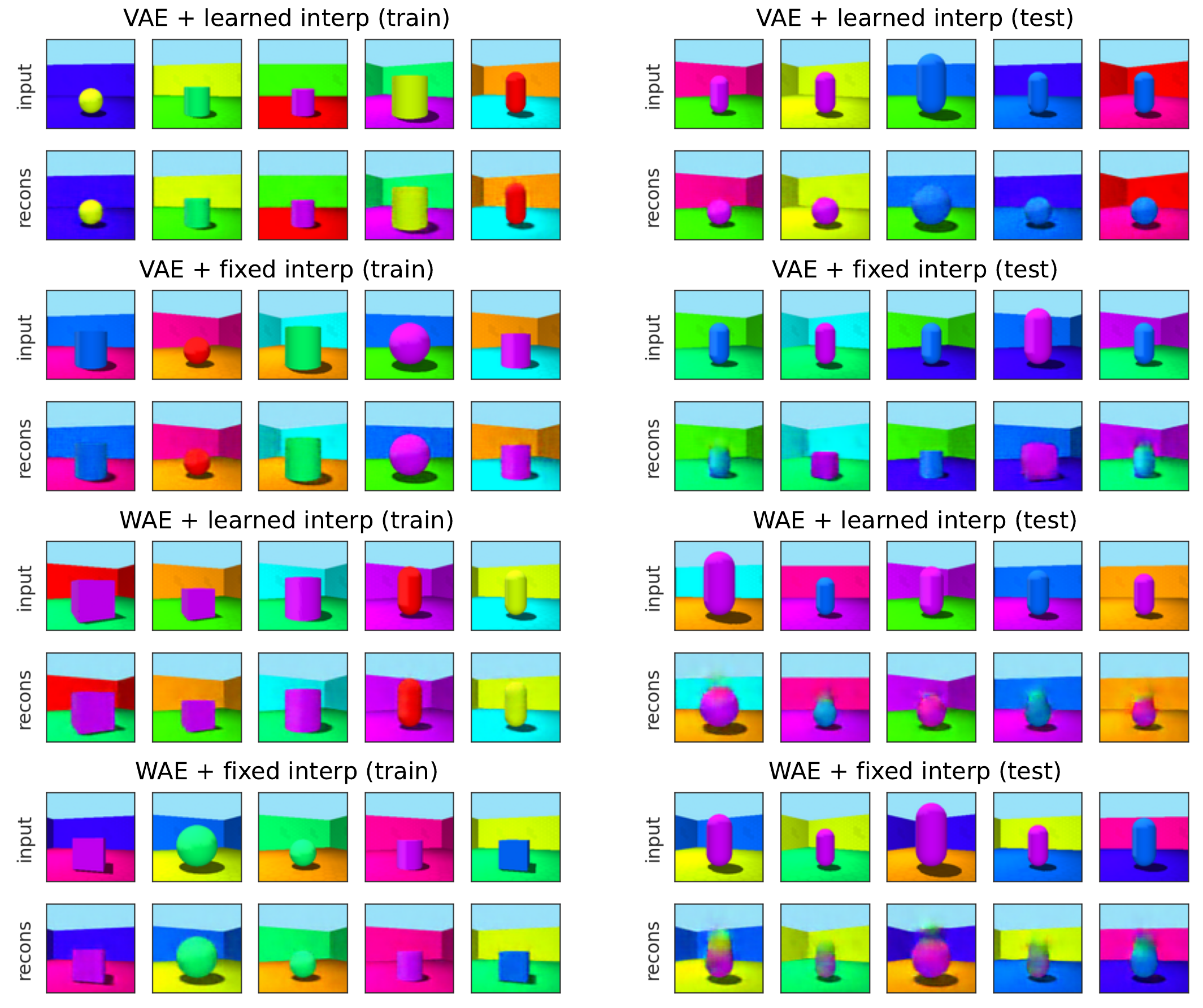}
    \caption{\textbf{Failures of combinatorial generalisation on \texttt{3DShapes}}Some example reconstructions for the \texttt{3DShapes} dataset for four models: (i) a VAE using learned interpolation (see Equations~\ref{eq:clearn}), (ii) a VAE using a fixed interpolation (see Equation~\ref{eq:caction}), (iii) a WAE using a learned interpolation, and (iv) a WAE using a fixed interpolation. All models learn representations that are highly disentangled. In each case, the five figures on the left show input (first row) and reconstruction (second row) for five training images. The five figures on the right show input and reconstructions for test images. These test images presented a novel combination of generative factors, here \comb{shape=pill, object hue~$>0.5$} -- that is, the model has seen all the combinations of shapes and hues for the object in the middle of the image, except for the `pill' shape, which has only been seen in ``warmer'' colors (object hue~$<0.5$). We observed that while reconstructions on the training dataset were really accurate - it is clear to see that the models struggled to reconstruct the test images, where the object was a `pill' shape with a ``cooler'' color.}
    \label{fig:shape2ohue-even-hues-recons}
\end{figure}

\begin{figure}[h!]
    \centering
    \includegraphics[width=\textwidth]{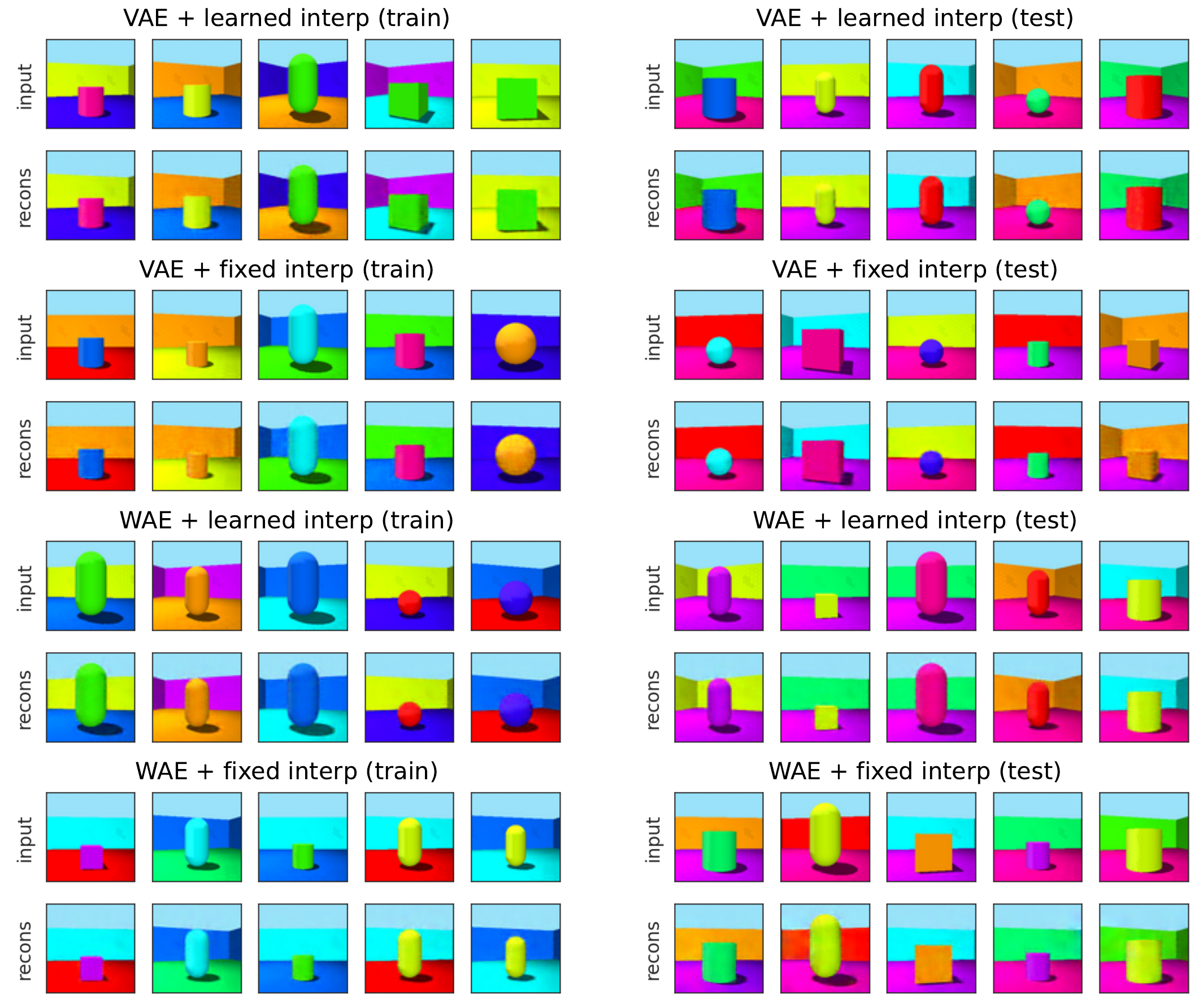}
    \caption{\textbf{Success in combinatorial generalisation on \texttt{3DShapes}} Some example reconstructions for the \texttt{3DShapes} dataset for four models: (i) a VAE using learned interpolation (see Equations~\ref{eq:clearn}), (ii) a VAE using a fixed interpolation (see Equation~\ref{eq:caction}), (iii) a WAE using a learned interpolation, and (iv) a WAE using a fixed interpolation. All models learn representations that are highly disentangled. In each case, the five figures on the left show input (first row) and reconstruction (second row) for five training images. The five figures on the right show input and reconstructions for test images. These test images presented a novel combination of generative factors, here \comb{floor hue~$<0.25$, wall hue~$>0.75$} -- that is, the model has seen all wall hues and floor hues in the range $[0, 1]$, but it has never seen a combination a floor with a hue~$<0.25$ with a wall of a hue~$>0.75$. We observed that, in this case, the model succeeded at combinatorial generalisation, reconstructing the images equally well on the training and test sets (compare with Figure~\ref{fig:shape2ohue-even-hues-recons}).}
    \label{fig:fhue2whue-recons}
\end{figure}

\begin{figure}[h!]
    \centering
    \includegraphics[width=\textwidth]{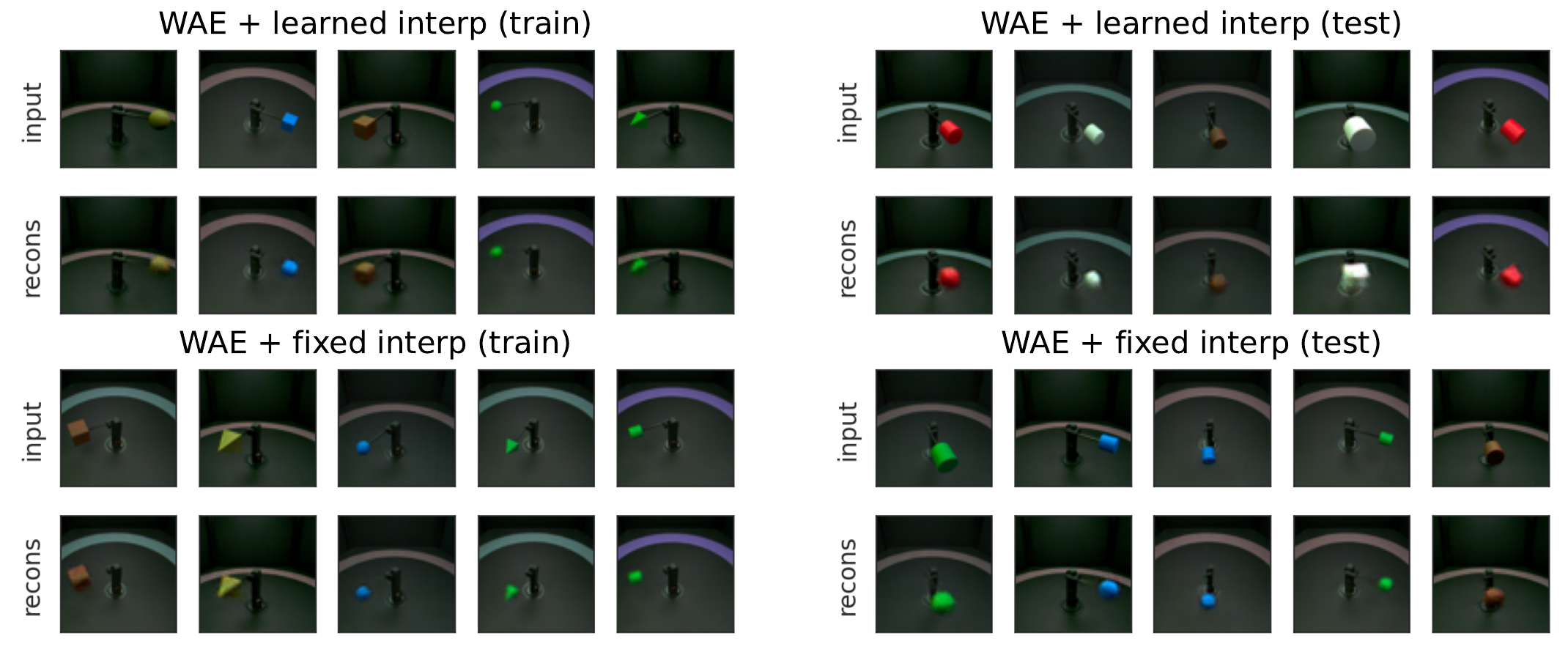}
    \caption{\textbf{Failures of combinatorial generalisation on \texttt{MPI3D}} Some example reconstructions for the \texttt{MPI3D} dataset for two models: (i) a WAE using learned interpolation (see Equations~\ref{eq:clearn}), (ii) a WAE using a fixed interpolation (see Equation~\ref{eq:caction}). Both models learn representations that are highly disentangled. In each case, the five figures on the left show input (first row) and reconstruction (second row) for five training images. The five figures on the right show input and reconstructions for test images. These test images presented a novel combination of generative factors, here \comb{shape=cylinder, vertical axis~$>0.5$} -- that is, the model has seen all the combinations of shapes (of the object at the end of a rod) and positions on the vertical axis, except for the cylinder, which has only been seen at vertical axis positions $<0.5$). We observed that while reconstructions on the training dataset were really accurate - both models struggled to reconstruct the test images with the left out combination of shape and position, frequently replacing the shape of the object.}
    \label{fig:shape2vc-recons}
\end{figure}

\begin{figure}[h!]
    \centering
    \includegraphics[width=\textwidth]{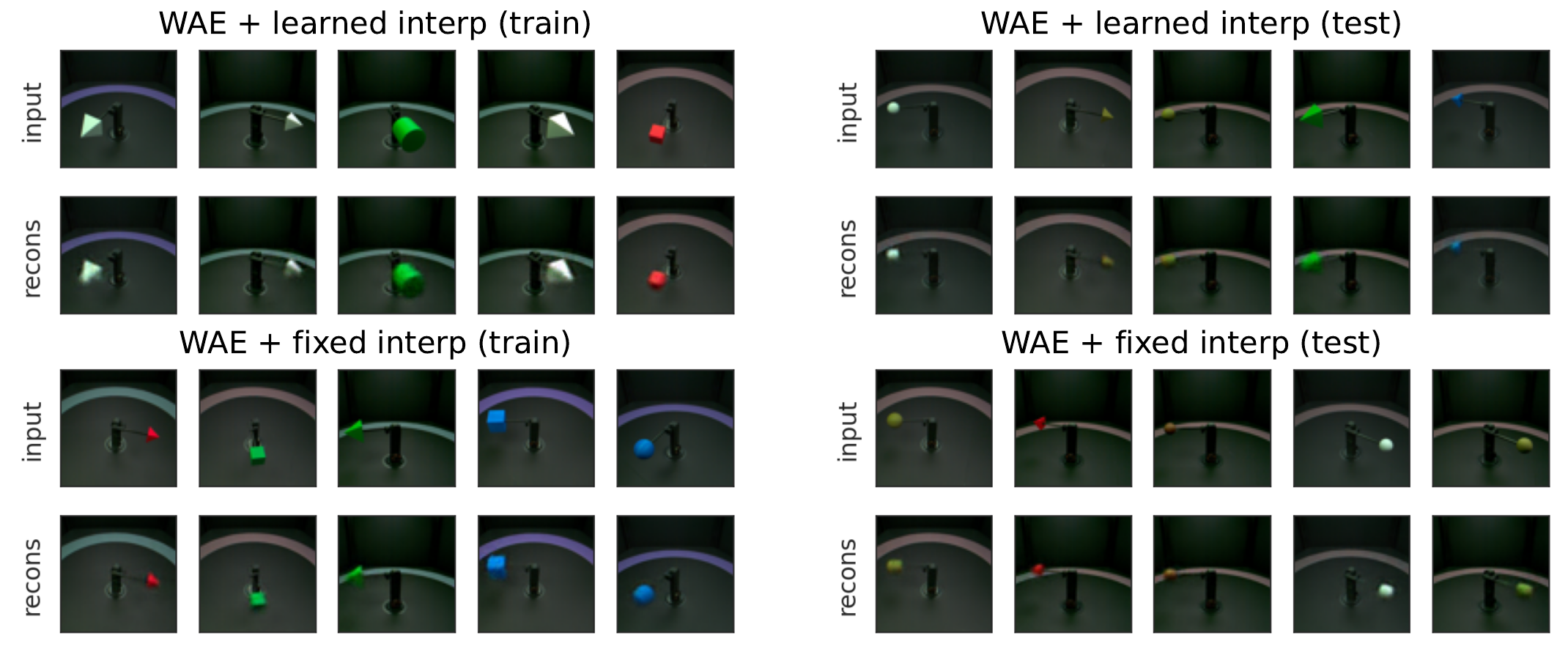}
    \caption{\textbf{Success in combinatorial generalisation on \texttt{MPI3D}} Some example reconstructions for the \texttt{MPI3D} dataset for two models: (i) a WAE using learned interpolation (see Equations~\ref{eq:clearn}), (ii) a WAE using a fixed interpolation (see Equation~\ref{eq:caction}). Both models learn representations that are highly disentangled. In each case, the five figures on the left show input (first row) and reconstruction (second row) for five training images. The five figures on the right show input and reconstructions for test images. These test images presented a novel combination of generative factors, here \comb{shape=\{cylinder,sphere\}, background color=salmon}. We observed that, in this case, the model managed to successfully perform the task of combinatorial generalisation, reproducing the training as well as the test images with really good accuracy (compare with \ref{fig:shape2vc-recons}).}
    \label{fig:shape2bkg-recons}
\end{figure}

\begin{figure}[h!]
    \centering
    \includegraphics[width=\textwidth]{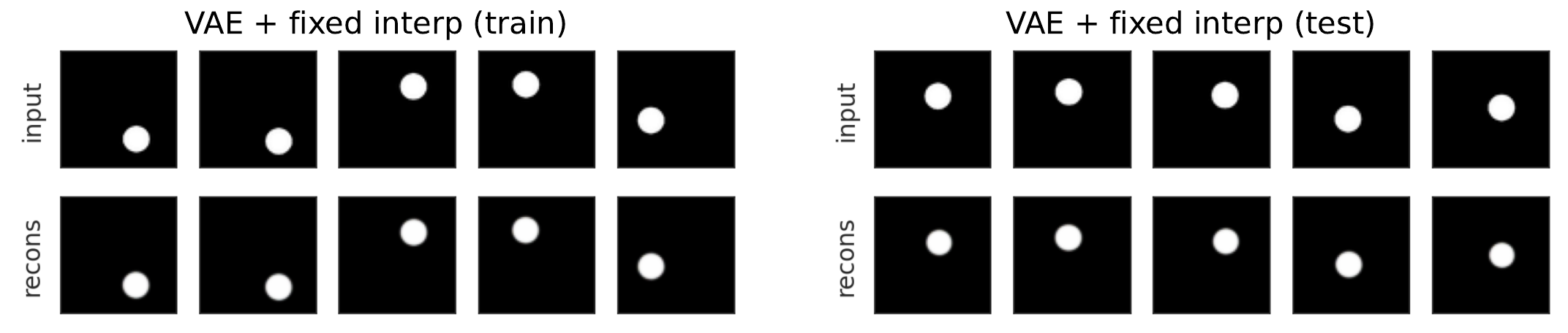}
    \caption{\textbf{Success in combinatorial generalisation on \texttt{Circles} dataset} Some example reconstructions for the \texttt{Circles} dataset. The model learns representations that are highly disentangled using fixed interpolation (see Equation~\ref{eq:caction}). The five figures on the left show input (first row) and reconstruction (second row) for five training images. The five figures on the right show input and reconstructions for test images. These test images presented a novel combination of generative factors, here \comb{$0.35<$~posX~$<0.65$, $0.35<$~posY~$<0.65$} -- that is, the model has seen circles in all x-positions, as well as all y-positions, but never for a combination of x \& y positions that fall in a central patch. We observed that the models managed to successfully perform the task of combinatorial generalisation, reproducing the circle in all positions of the canvas.}
    \label{fig:circles-midpos-recons}
\end{figure}

\begin{figure}[h!]
    \centering
    \includegraphics[width=\textwidth]{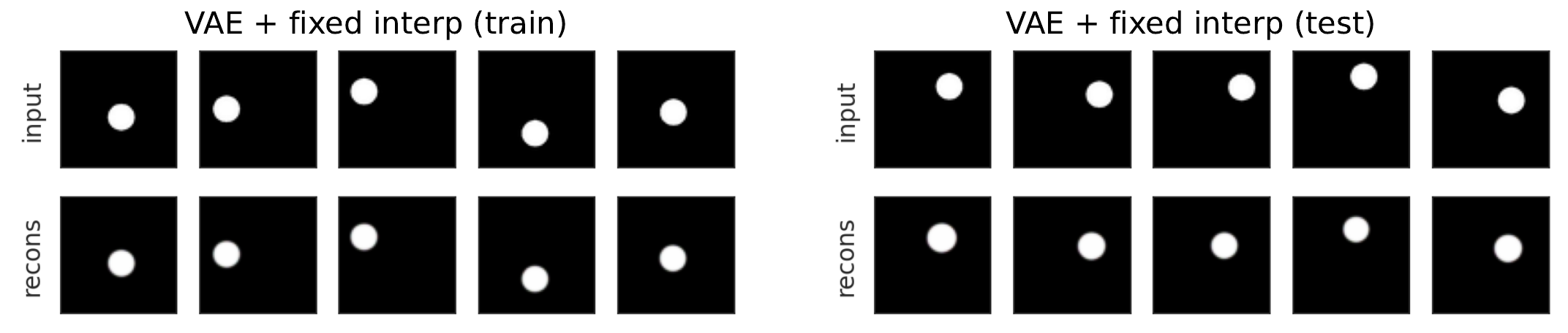}
    \caption{\textbf{Failure in combinatorial generalisation on \texttt{Circles} dataset} Some example reconstructions for the \texttt{Circles} dataset. The model learns representations that are highly disentangled using fixed interpolation (see Equation~\ref{eq:caction}). The five figures on the left show input (first row) and reconstruction (second row) for five training images. The five figures on the right show input and reconstructions for test images. These test images presented a novel combination of generative factors, here \comb{posX~$>0.5$, posY~$>0.5$} -- that is, the model has seen circles in all x-positions, as well as all y-positions, but never for a combination of x \& y positions that fall in a top right corner. We observed that the models failed in this condition in line with results in \citet{watters_spatial_2019}}
    \label{fig:circles-corner-recons}
\end{figure}

\begin{figure}[h!]
    \centering
    \includegraphics[width=\textwidth]{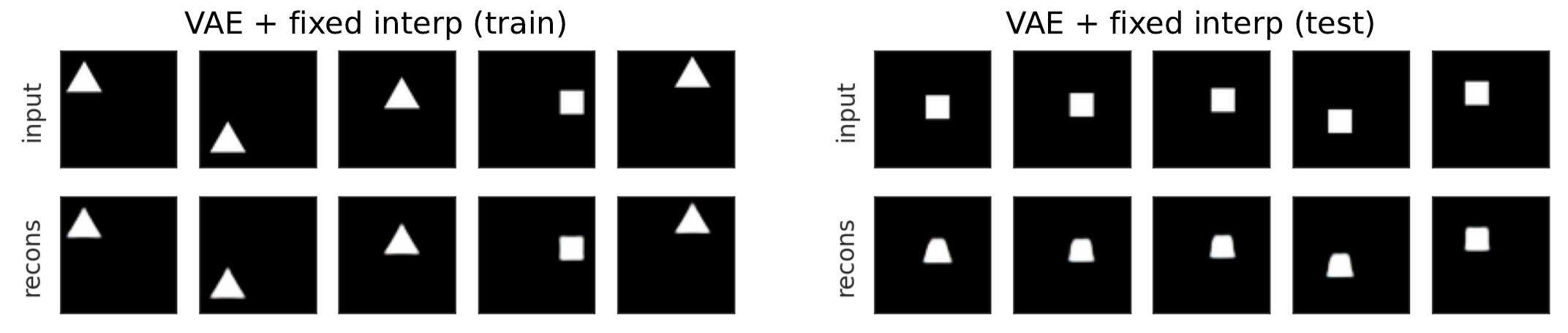}
    \caption{\textbf{Failures of combinatorial generalisation on \texttt{Simple} dataset for the middle patch condition} Some example reconstructions for the \texttt{Simple} dataset for two models: The model learns representations that are highly disentangled using fixed interpolation (see Equation~\ref{eq:caction}). The five figures on the left show input (first row) and reconstruction (second row) for five training images. The five figures on the right show input and reconstructions for test images. These test images presented a novel combination of generative factors, here \comb{$0.35<$~posX~$<0.65$, $0.35<$~posY~$<0.65$,shape=triangle} -- that is, the model has seen triangles in all x-positions, as well as all y-positions, but never for a combination of x \& y positions that fall in a central patch. It has seen squares in a central location. We observed that the models failed to proprely generalise in this condition, as opposed to the corresponding condition in the circles dataset (see Figure~\ref{fig:circles-midpos-recons}}
    \label{fig:simple-recons}
\end{figure}

\FloatBarrier
\newpage
\subsection{Disentanglement}\label{app:disent}

\begin{figure}[h!]
    \centering
    \includegraphics[width=\textwidth]{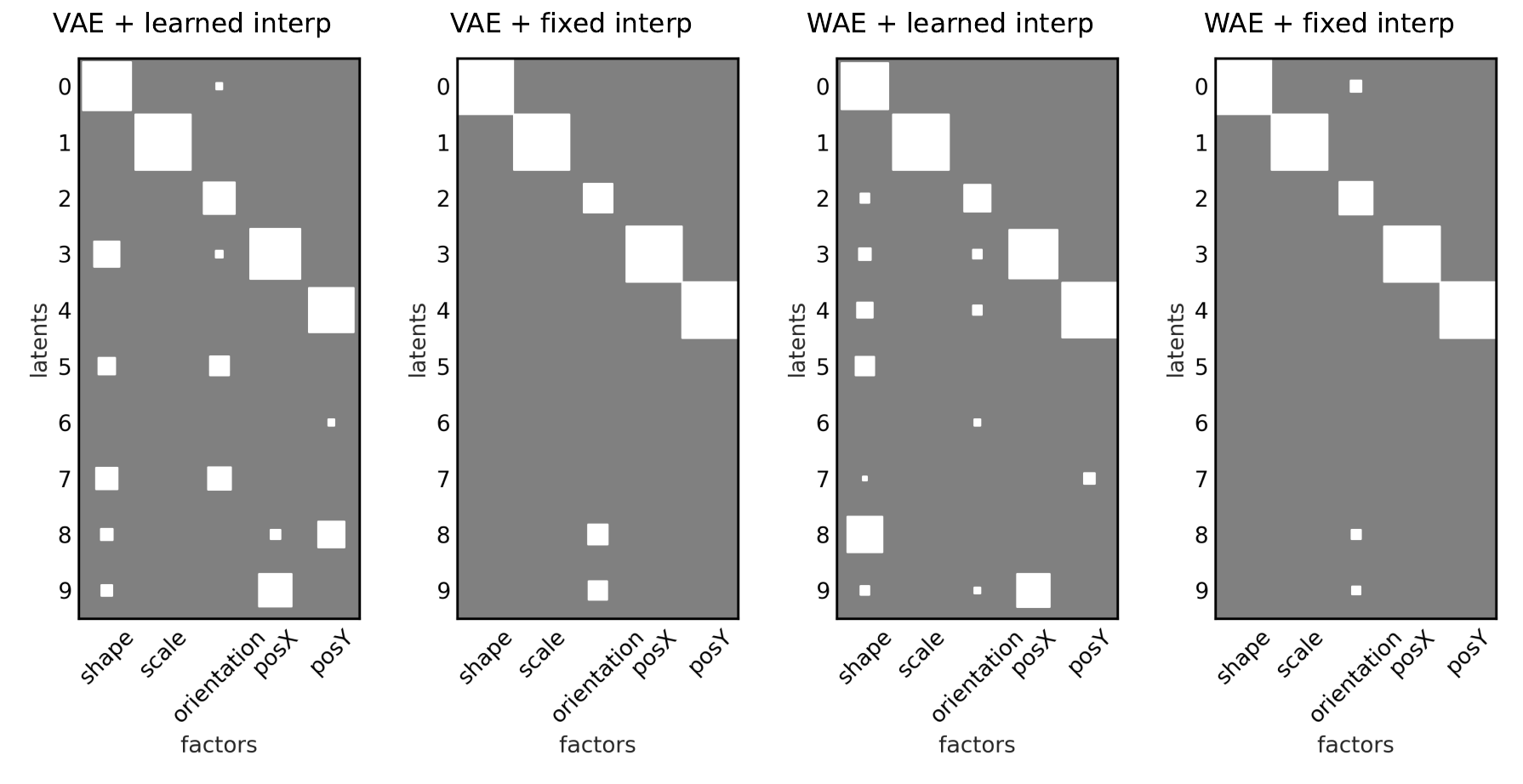}
    \caption{\textbf{Visualising disentanglement for \texttt{dSprites} dataset} Following \citet{eastwood_framework_2018}, we used Hinton matrices to visualise the extent of disentanglement. Each panel shows the matrix $C$ (see Section~\ref{app:disent-def}) of regression coefficients between latent variables and generative factors. A disentangled model should have high coefficients for each pair of generative factors and latent variable but low coefficients everywhere else -- i.e. the matrix should be sparse (see ideal Hinton matrix in Figure~\ref{fig:ideal-disent}). Here we can see that all models learning the composition task show highly sparse matrices and models using fixed interpolation (Equation~\ref{eq:caction}) achieve particularly high disentanglement.}
    \label{fig:disent-sqr2px-hinton}
\end{figure}

\begin{figure}[h!]
    \centering
    \includegraphics[width=\textwidth]{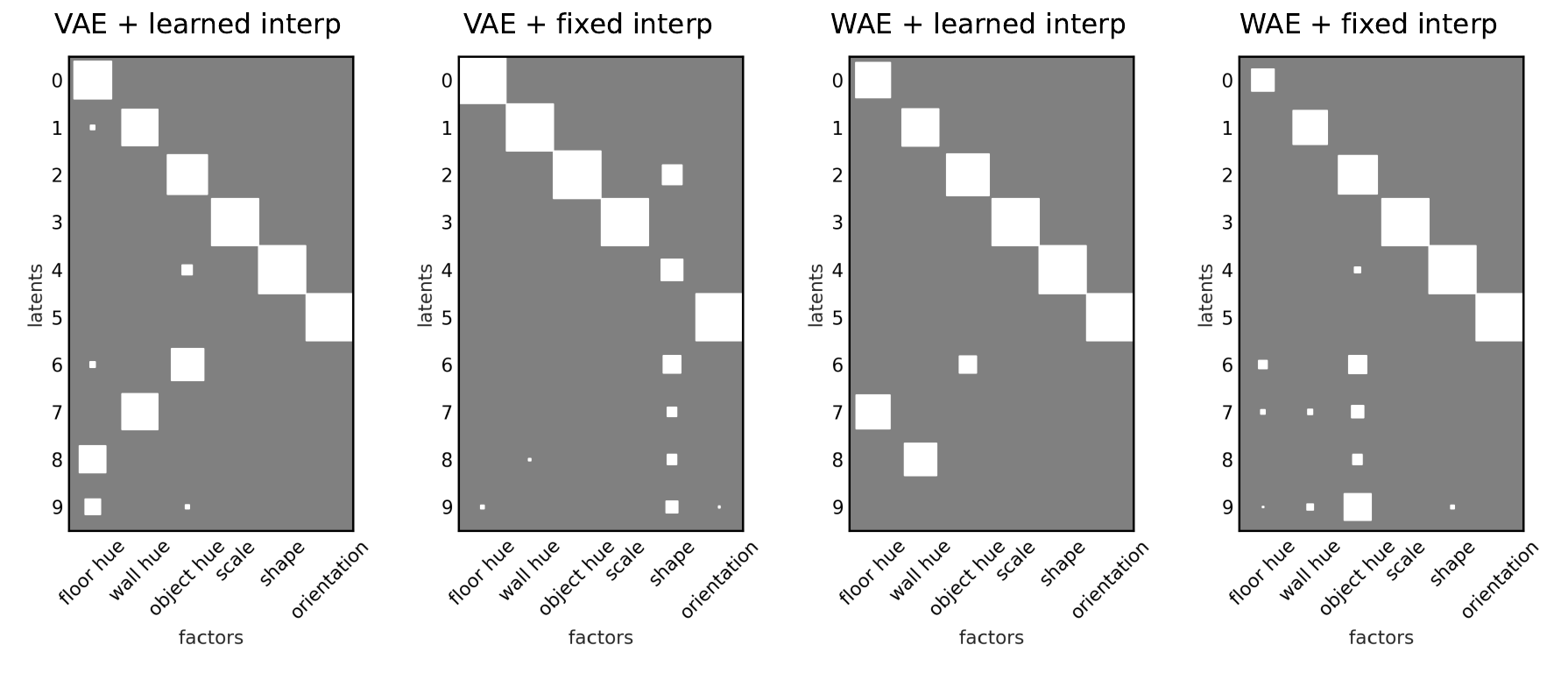}
    \caption{\textbf{Visualising disentanglement for \texttt{3DShapes} dataset, failure condition} Following \citet{eastwood_framework_2018}, we used Hinton matrices to visualise the extent of disentanglement. Each panel shows the matrix $C$ (see Section~\ref{app:disent-def}) of regression coefficients between latent variables and generative factors.}
    \label{fig:shape2ohue-even-hues-hinton}
\end{figure}

\begin{figure}[h!]
    \centering
    \includegraphics[width=\textwidth]{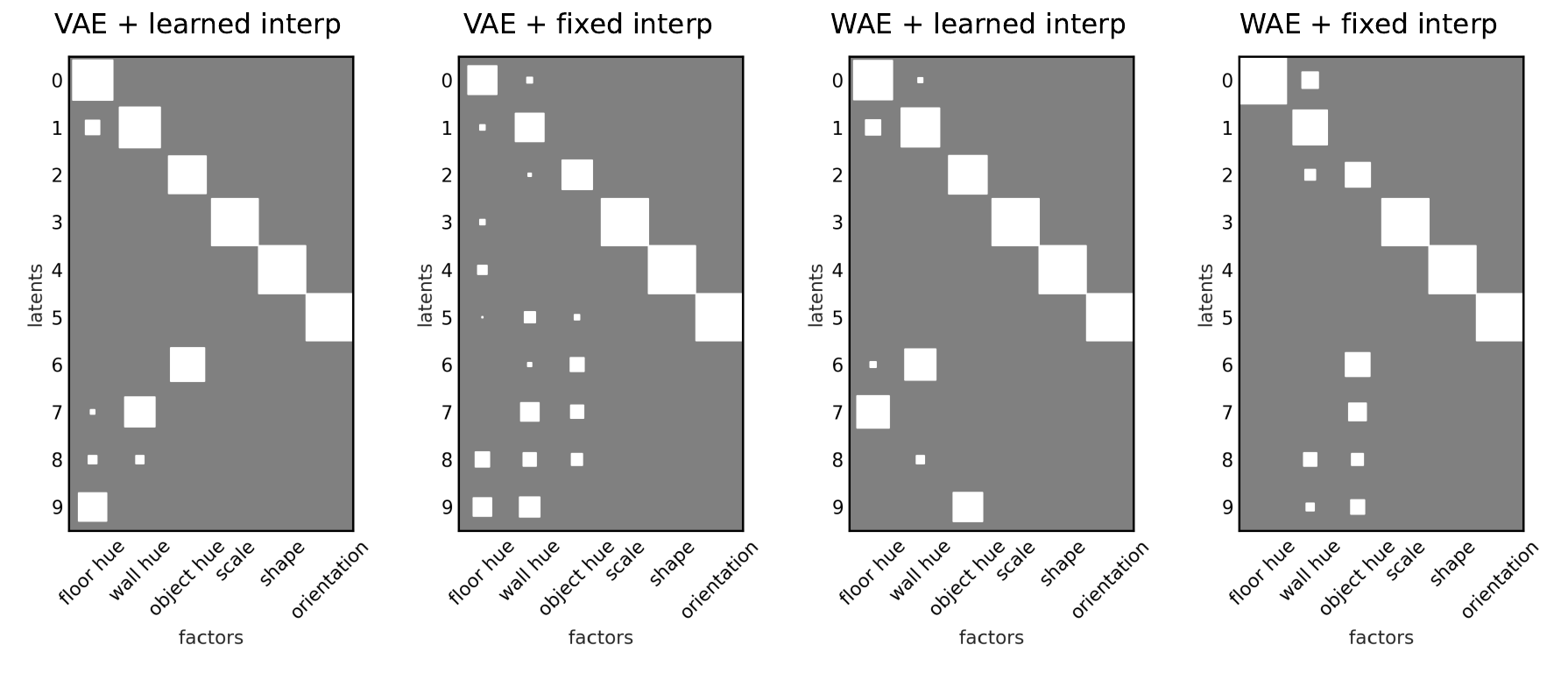}
    \caption{\textbf{Visualising disentanglement for \texttt{3DShapes} dataset, success condition} Following \citet{eastwood_framework_2018}, we used Hinton matrices to visualise the extent of disentanglement. Each panel shows the matrix $C$ (see Section~\ref{app:disent-def}) of regression coefficients between latent variables and generative factors.}
    \label{fig:fhue2whue-hinton}
\end{figure}

\begin{figure}[h!]
    \centering
    \includegraphics[width=0.5\textwidth]{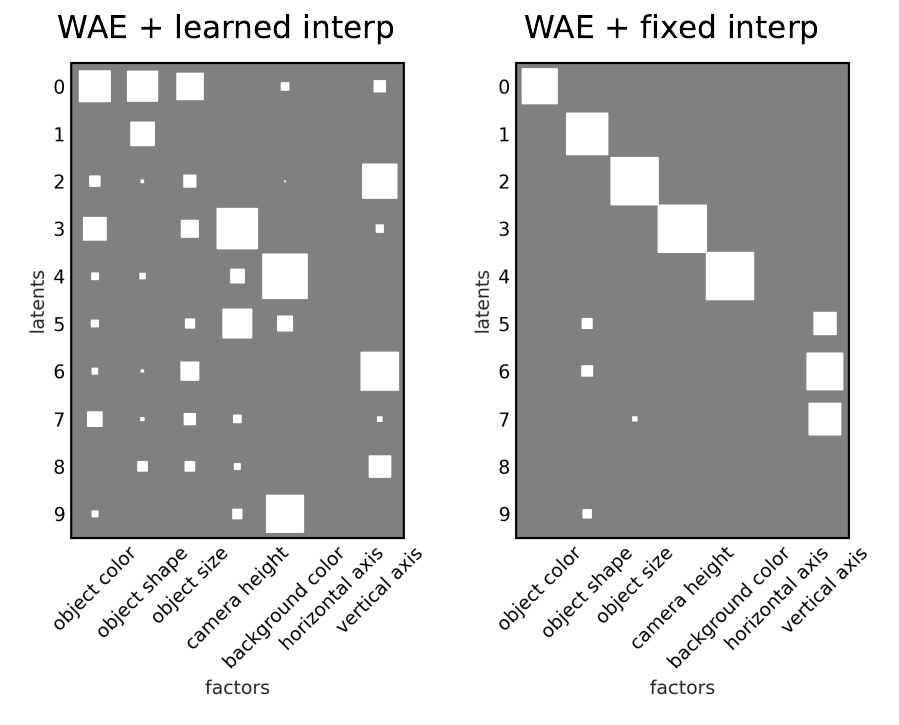}
    \caption{\textbf{Visualising disentanglement for \texttt{MPI3D} dataset, failure condition} Following \citet{eastwood_framework_2018}, we used Hinton matrices to visualise the extent of disentanglement. Each panel shows the matrix $C$ (see Section~\ref{app:disent-def}) of regression coefficients between latent variables and generative factors.}
    \label{fig:shape2vc-hinton}
\end{figure}

\begin{figure}[h!]
    \centering
    \includegraphics[width=0.5\textwidth]{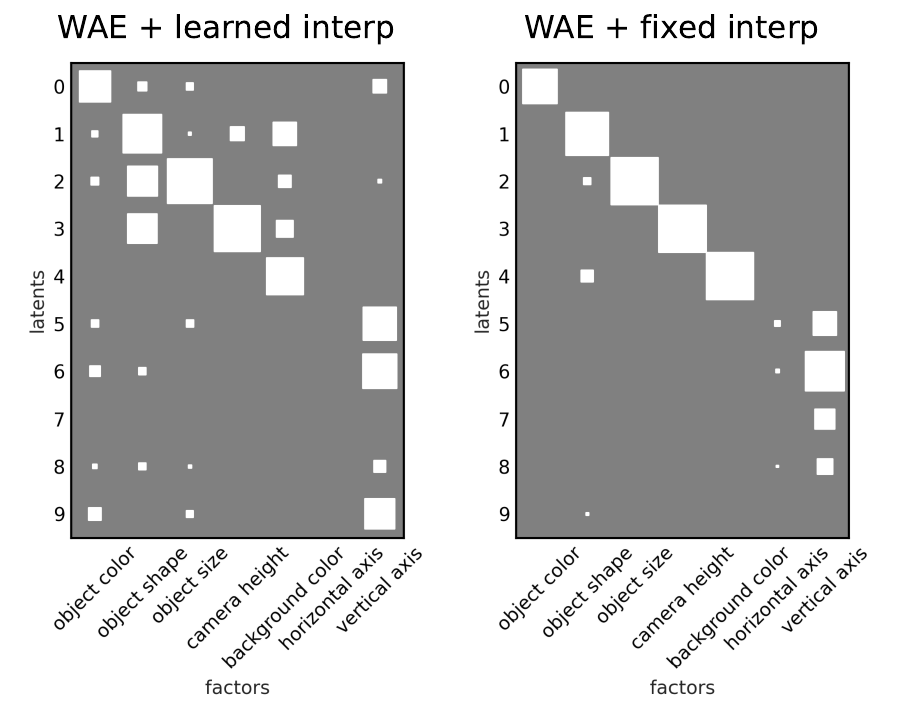}
    \caption{\textbf{Visualising disentanglement for \texttt{MPI3D} dataset, success condition} Following \citet{eastwood_framework_2018}, we used Hinton matrices to visualise the extent of disentanglement. Each panel shows the matrix $C$ (see Section~\ref{app:disent-def}) of regression coefficients between latent variables and generative factors.}
    \label{fig:shape2bkg-hinton}
\end{figure}

\begin{figure}[h!]
    \centering
    \begin{subfigure}{0.4\textwidth}
        \centering
        \includegraphics[width=0.9\linewidth]{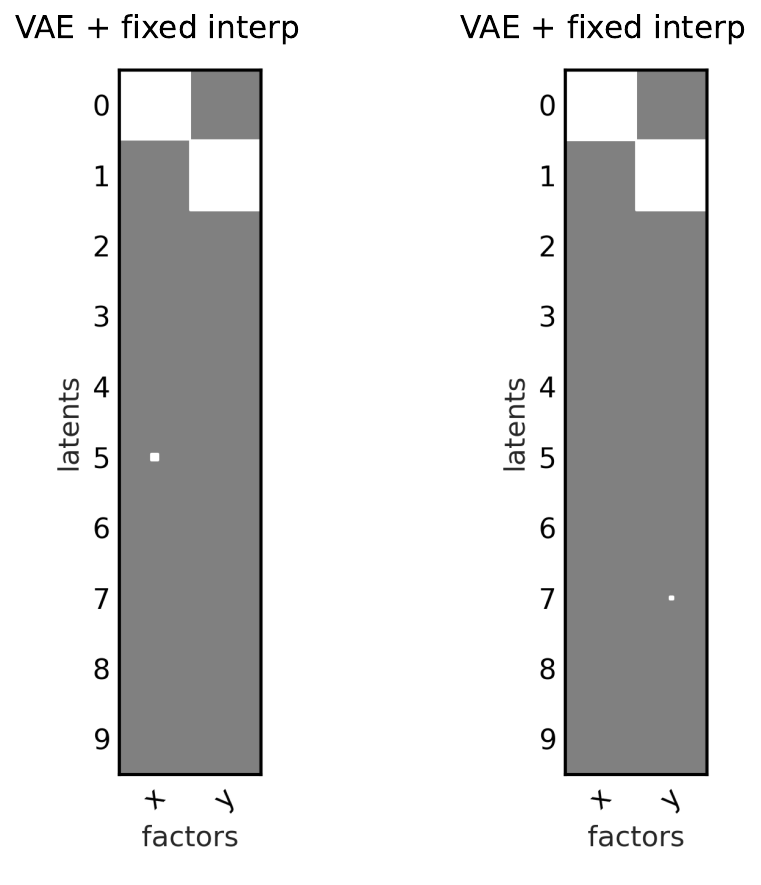}
        \caption{\textbf{\texttt{Circles} dataset} Following \citet{eastwood_framework_2018}, we used Hinton matrices to visualise the extent of disentanglement. Each panel shows the matrix $C$ (see Section~\ref{app:disent-def}) of regression coefficients between latent variables and generative factors. \textit{Left} the corner condition. \textit{The middle patch condition}.}
        \label{fig:circles-hinton}
    \end{subfigure}%
    \hspace{1em}
    \begin{subfigure}{0.4\textwidth}
        \centering
        \includegraphics[width=0.9\linewidth]{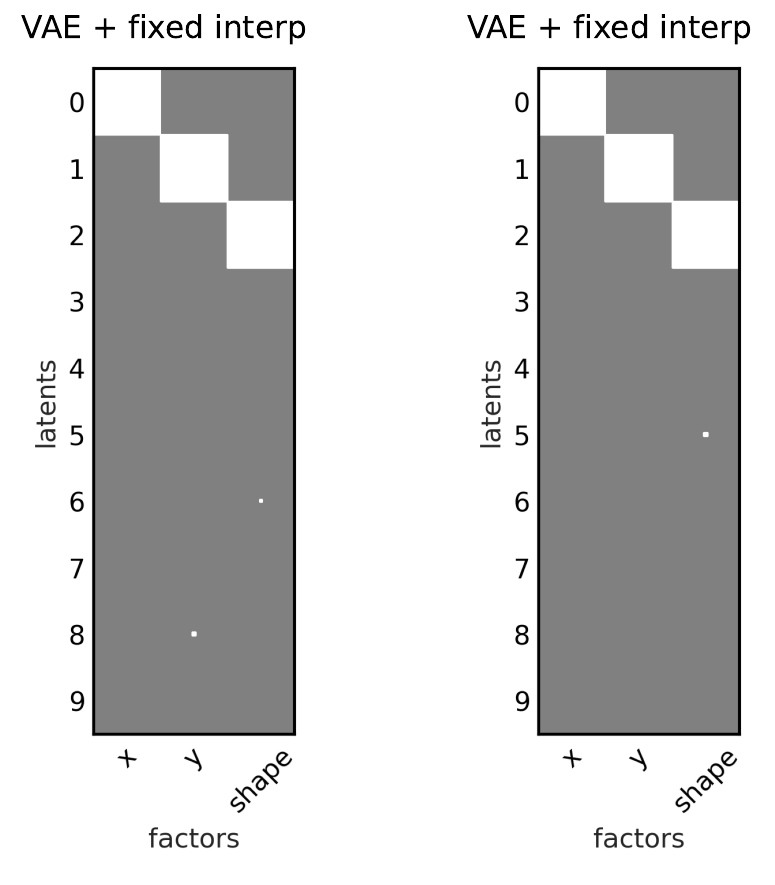}
        \caption{\textbf{\texttt{Simple} dataset} Following \citet{eastwood_framework_2018}, we used Hinton matrices to visualise the extent of disentanglement. Each panel shows the matrix $C$ (see Section~\ref{app:disent-def}) of regression coefficients between latent variables and generative factors. \textit{Left} the corner condition. \textit{The middle patch condition}.}
        \label{fig:simple-hinton}
    \end{subfigure}
    \caption{\textbf{Visualising disentanglement for the \texttt{Circles} and \texttt{Simple} datasets}}
\end{figure}

\FloatBarrier
\newpage
\subsection{Latent visualizations}

\begin{figure}[h!]
    \centering
    \includegraphics[width=\textwidth]{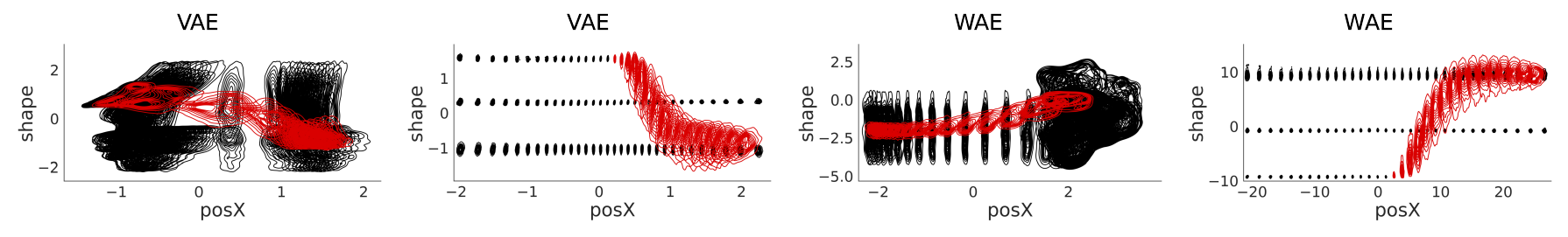}
    \caption{\textbf{Visualisation of latent space for \texttt{dSprites} dataset}. Distribution of encoded values for different combinations of shape and posX for the dSprites dataset. For models that do not disentangle well (first and third from left to right), we observe a very unstructured distribution of both training (black) and (test). Models with high disentanglement (second and fourth) show a very structured representation where it is easy to discriminate between different shapes and positions in the training set, but which diverge in the test set.}
    \label{fig:disent-sqr2px-latent}
\end{figure}

\begin{figure}[h!]
    \centering
    \includegraphics[width=\textwidth]{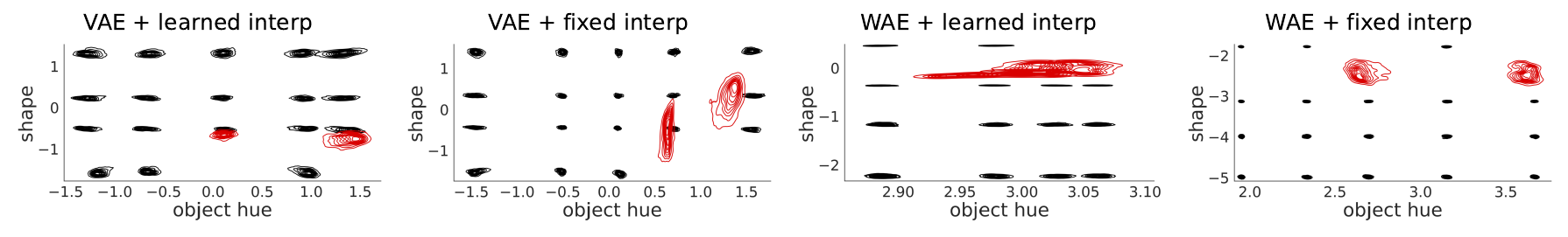}
    \caption{\textbf{Visualisation of latent space for \texttt{3DShapes} dataset, failure condition}. Distribution of encoded values for different combinations of shape and object hue for the 3DShapes dataset. In this case, all models showed high levels of disentanglement. The factor combinations for the training set are easily distinguishable. However, models show poor generalisation in the test set. The 4th model on the right is the one that shows the best results, yet the test distributions show a signfican increase in variance and drift towards observed training data.}
    \label{fig:shape2ohue-even-hues-latent}
\end{figure}

\begin{figure}[h!]
    \centering
    \includegraphics[width=\textwidth]{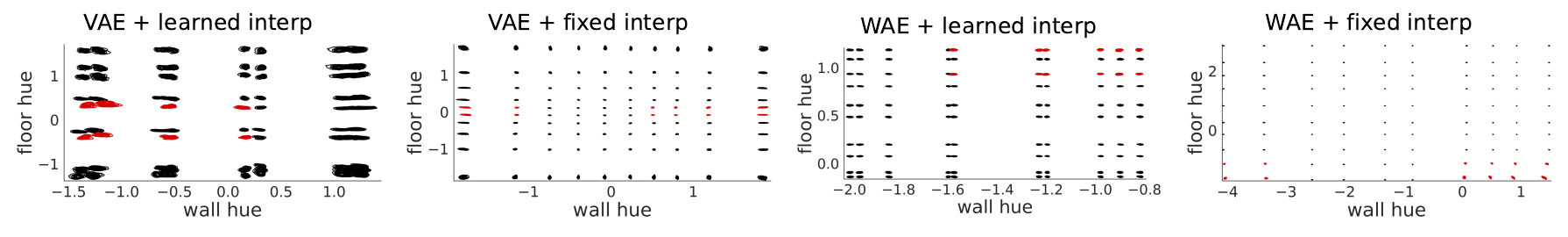}
    \caption{\textbf{Visualisation of latent space for \texttt{3DShapes} dataset, success condition}. Distribution of encoded values for different combinations of wall hue and floor hue for the 3DShapes dataset. In this case, all models showed high levels of disentanglement. The factor combinations for the training set are easily distinguishable. As opposed to the previous condition, models show excellent generalisation to the test data.}
    \label{fig:fhue2whue-latent}
\end{figure}

\begin{figure}[h!]
    \centering
    \includegraphics[width=0.65\textwidth]{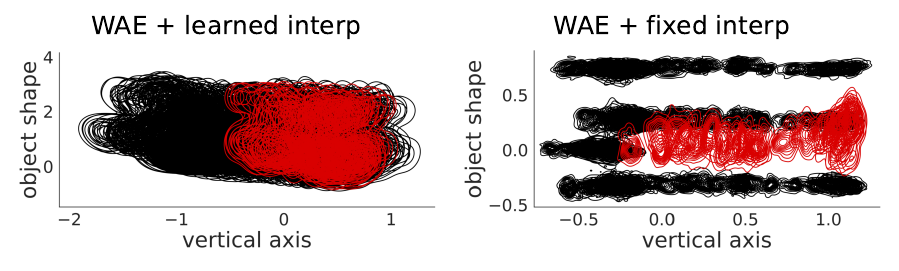}
    \caption{\textbf{Visualisation of latent space for \texttt{MPI3D} dataset, failure condition}. Distribution of encoded values for different combinations of shape and vertical axis for the MPI3D dataset. The model on the left shows poor disentanglement with the joint distribution of the best match latents for both training (black) and testing examples (red) showing significant overlap. The model on the right on the other hand achieved high disentanglement. The encoded value for each shape is easily distinguishable. The encoded angle of the arm is not as clearly separated. The test data shows a significant drift, though not as dramatic as in \texttt{dSprites}.}
    \label{fig:shape2vc-latent}
\end{figure}

\begin{figure}[h!]
    \centering
    \includegraphics[width=0.65\textwidth]{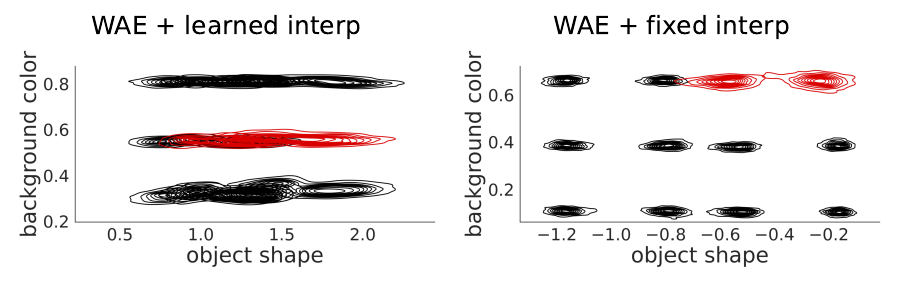}
    \caption{\textbf{Visualisation of latent space for \texttt{MPI3D} dataset, success condition}. Distribution of encoded values for different combinations of shape and vertical axis for the MPI3D dataset. The first three models from the left show poor disentanglement with the joint distribution for both training (black) and testing examples (red). The model on the right on the other hand achieved high disentanglement. The encoded value for each shape background color combination is easily distinguishable. There is no hint of drift or even increase in variance.}
    \label{fig:shape2bkg-latent}
\end{figure}

\begin{figure}[h!]
    \centering
    \includegraphics[width=0.65\textwidth]{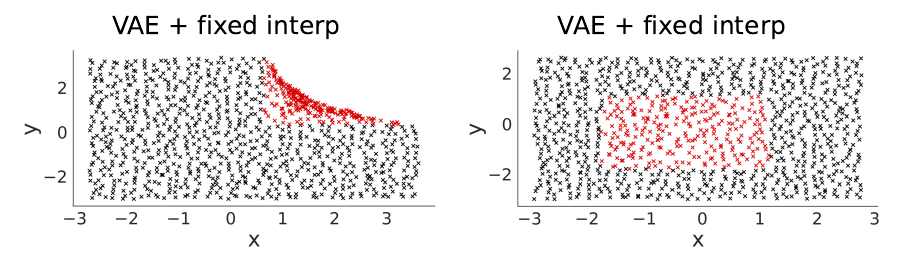}
    \caption{\textbf{Visualisation of latent space for \texttt{Circles} dataset}. We replicate the results from \citeauthor{watters_spatial_2019}. On the left, the Spatial Broadcast Decoder shows poor generalisation when tested on the corner case. Test examples (red) bend over the closest seen training examples (black). On the right, when the excluded patch is locate in the center, the model shows excellent generalisation.}
    \label{fig:circles-latent}
\end{figure}

\begin{figure}[h!]
    \centering
    \includegraphics[width=0.65\textwidth]{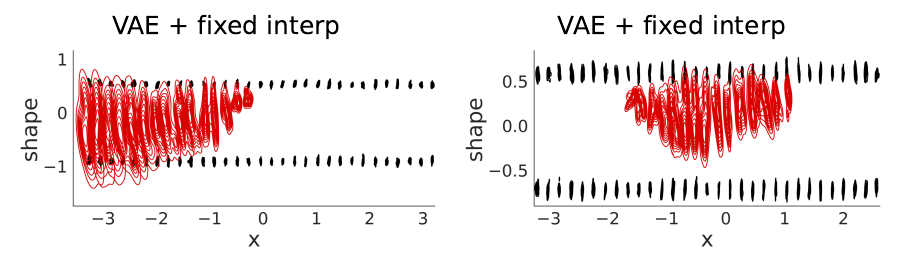}
    \caption{\textbf{Visualisation of latent space for \texttt{Simple} dataset}. We show that the Spatial Broadcast Decoder \citep{watters_spatial_2019} fails when we introduce another shape in the middle patch position. On the left, the result for the corner case, showing again poor generalisation. The drift is similar to the one observed in dSprites: as we move away from observed positions the latent representation experience a dramatic increase in variance and shift towards the mean of the other shape. On the right, when the excluded patch is locate in the center, the model also fails as the introduction of a new shape makes turns the task too hard for the model.}
    \label{fig:simple-latent}
\end{figure}

\FloatBarrier

\clearpage
\raggedbottom
\section{Interactive vs Non-interactive combinations of generative factors}

In our simulations, we observed that models failed at combinatorial generalisation when the generative factor `shape' was combined with another generative factor (position / color). We hypothesised that this was because shape combines with other factors (position / color / orientation) in an \emph{interactive} manner -- that is, the value of any pixel in the image is determined jointly by the value of \comb{shape} and other factors. To understand why this is a hard problem, consider a graphical representation that captures the dependencies between the pixels of an image and generative factors (Figure~\ref{fig:interactive}.a). Each pixel and each generative factor corresponds to a node in this graph -- pixel nodes are at the bottom and generative factor nodes are at the top. The value of the pixel nodes depends on the value of one or more generative factors. We capture this dependence as edges between pixel nodes and generative factor nodes. 

Consider two different conditions. In the first condition, we take a combination of the generative factors \comb{floor-hue, wall-hue}. In this case, the factor node \comb{floor-hue} will determine one set of pixel nodes, while the factor \comb{wall-hue} will determine another (mutually-exclusive) set of pixel nodes. During training, the model needs to learn how to map the value of each generative factor to the corresponding pixel. Once it has learned this mapping, it can easily generalise to unseen combinations of \comb{floor-hue, wall-hue} because the same graphical dependencies learned from training work. We call this case the \emph{non-interactive} condition.

Now, consider a second condition where the generative factors are \comb{shape, posX}. In this case, the value of each pixel node is \emph{jointly} determined by the values of both generative factors. Furthermore, note that the mapping is highly nonlinear and complex. For example, in the dSprites dataset, a pixel node may have value $+1$ for a given value of \comb{shape} and \comb{posX}, but this value may change to $-1$ for a slight change in either factor. Other changes will have no effect on the value of some pixels. To succeed at combinatorial generalisation, a model must account for these dependencies and any changes that occur dynamically as a result of the interaction between the generative factors. We call this condition \emph{interactive} and hypothesised that models will succeed in the non-interactive condition, but succeed in the interactive condition. The results in Figures~\ref{fig:results-comp} and \ref{fig:success-results} confirm this hypothesis.

\FloatBarrier

\clearpage
\section{Control experiments}\label{app:controls}

\subsection{Rebalanced datsets}

In all experiments where we tested combinatorial generalisation, combinations along a subset of dimensions were excluded from the training set. This means that the model was trained on an unequal number of samples from different dimensions. Here we include results for simulations where the datasets have been rebalanced so that all shapes are seen the same number of times. Figures show that models achieve high levels of disentanglement while still failing to generalise. Since we observed that models achieve disentanglement more consistently compared to an unbalanced dataset, we rebalance the dataset in all subsequent experiments. 

\begin{figure}[h!]
    \centering
    \includegraphics[width=\textwidth]{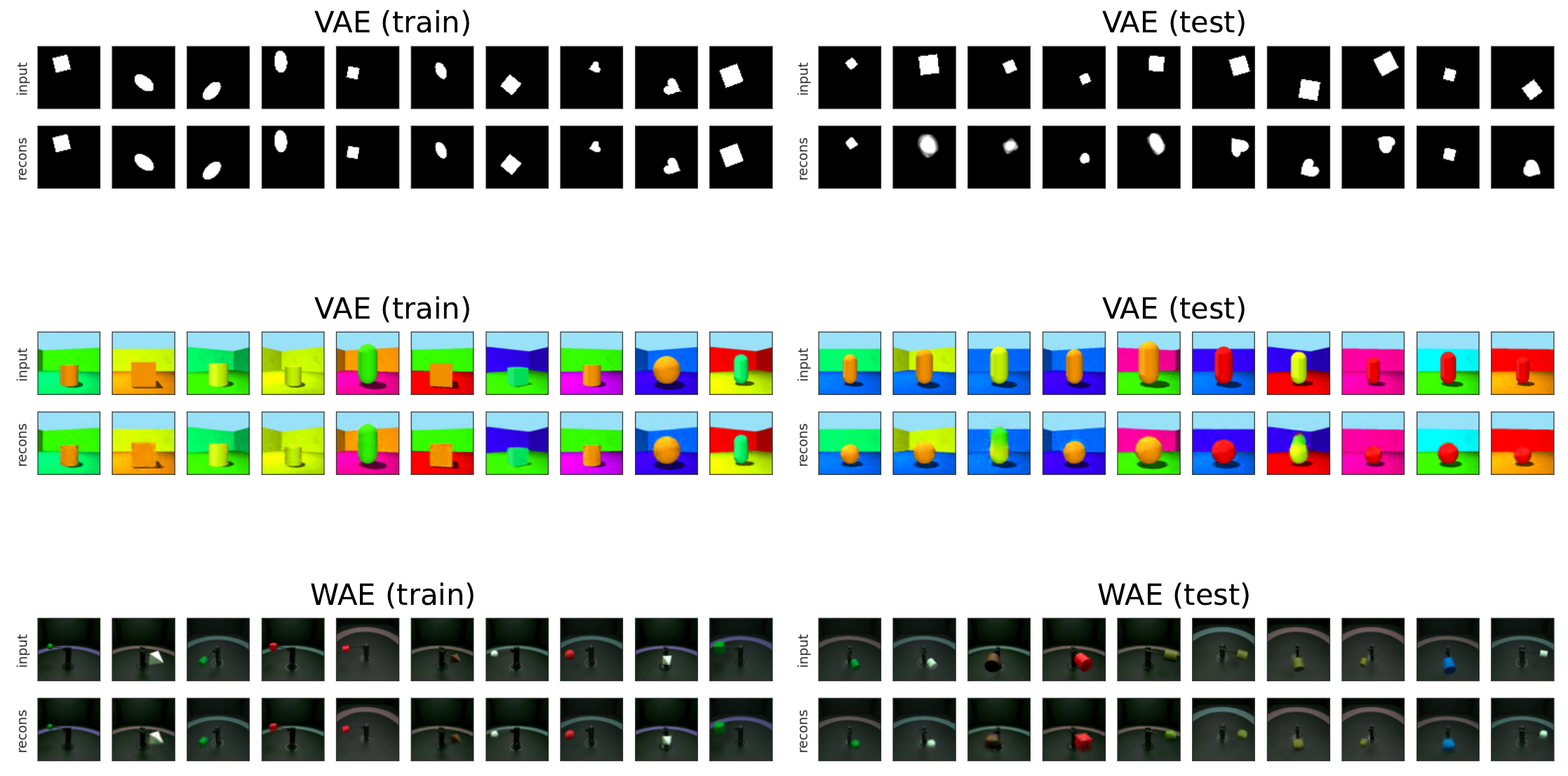}
    \caption{\textbf{Failures of combinatorial generalisation for models trained on dSprites, 3DShapes and MPI3d}. In each case, the top row shows input images and the bottom row shows reconstructions. The images on the left are sampled from the training set and the ones on the right from the test set. Failures of generalisation usually manifest as the model swapping the input shape. Only harder conditions with rebalanced datsets (i.e. all shapes are observed the same number of times) are shown.}
    \label{fig:rebalanced-hard}
\end{figure}

\begin{figure}[h!]
    \centering
    \includegraphics[width=\textwidth]{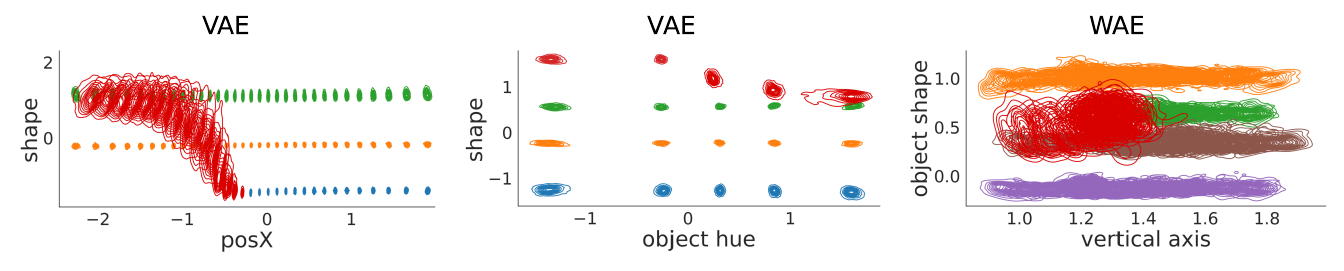}
    \caption{\textbf{Visualisation of latent space for models trained on dSprites, 3DShapes and MPI3d}. Like in other results, the latent space projections for test images are shown in red. We have also colour-coded the training set images according to shape. This makes it clear that the models are clearly disentangled (one shape (colour) maps to one row (value of latent variable)). Despite this, we see generalisation failures, where left-out combinations show an increase in variance and the mean drifts to values seen during the training set. Only harder conditions with rebalanced datsets (i.e. all shapes are observed the same number of times) are shown.}
    \label{fig:rebalance}
\end{figure}

\begin{figure}[h!]
    \centering
    \includegraphics[width=0.5\textwidth]{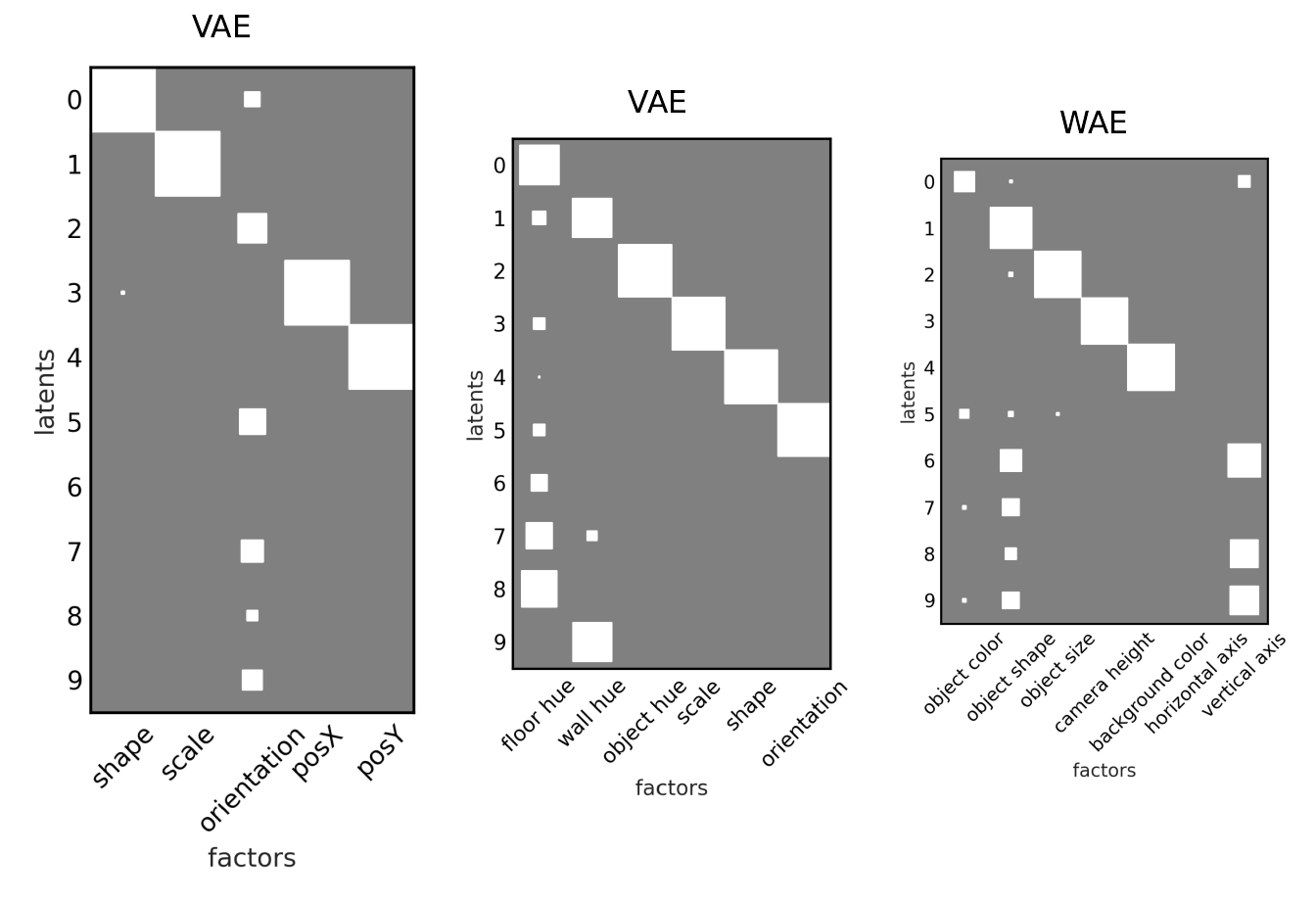}
    \caption{\textbf{Visualising disentanglement for models trained on dSprites, 3DShapes and MPI3d}. Only harder conditions with rebalanced datsets (i.e. all shapes are observed the same number of times) are shown.}
    \label{fig:shape2bkg-hinton-rebalanced}
\end{figure}

\FloatBarrier

\clearpage
\subsection{Using an ideal decoder}\label{app:frankenstein}

For all results in the main text where we use the semi-supervised (composition) task, the model was trained end-to-end. This means that the encoder and decoder were jointly trained. Then at test time, we found that the model failed to reconstruct images containing left-out combinations of generative factors and the encoder also failed to project these images with left-out combinations to the correct region of the latent space. However, it could be argued that the failure to project images in the correct region of latent space is not entirely down to the encoder as the encoder and decoder were jointly trained. In other words, if one could fix the decoder -- i.e. the decoder can learn to correctly map regions in the latent space to image space -- then the encoder may learn to perform combinatorial generalisation. To test this hypothesis we ran the dSprites task using an ``ideal decoder''.

\begin{figure}
    \centering
    \includegraphics[width=0.7\textwidth]{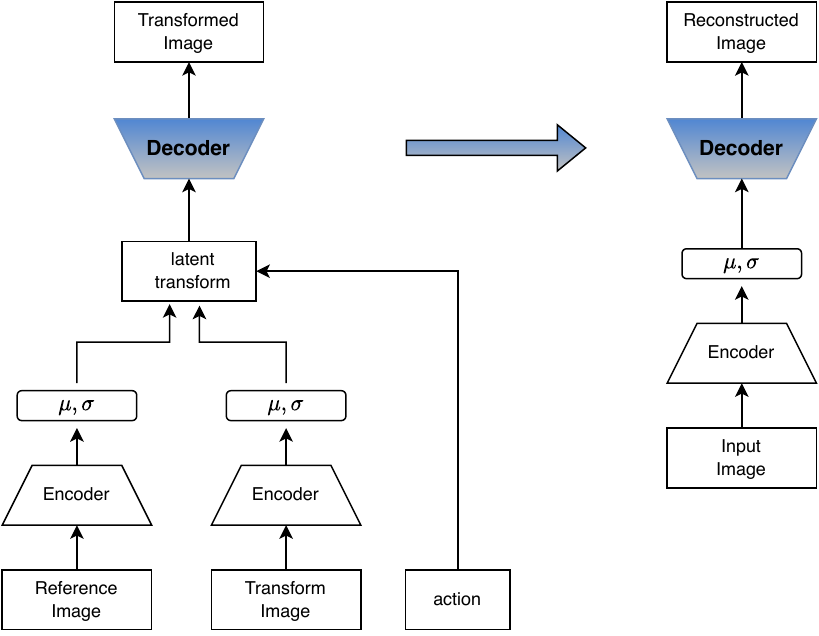}
    \caption{\textbf{Schematic illustration of the ideal decoder experiment.} The ``ideal decoder'' was constructed by training the model on the composition task (left) using the entire dataset (here dSprites) and then inserting the trained decoder from this setup into a VAE which was then trained on the image reconstruction task (right). All weights of the decoder were frozen, essentially limiting the learning to the encoder.}
    \label{fig:frankenstein-schema}
\end{figure}

\begin{figure}[h!]
    \centering 
    \begin{subfigure}{\textwidth}
        \centering
        \includegraphics[width=0.7\linewidth]{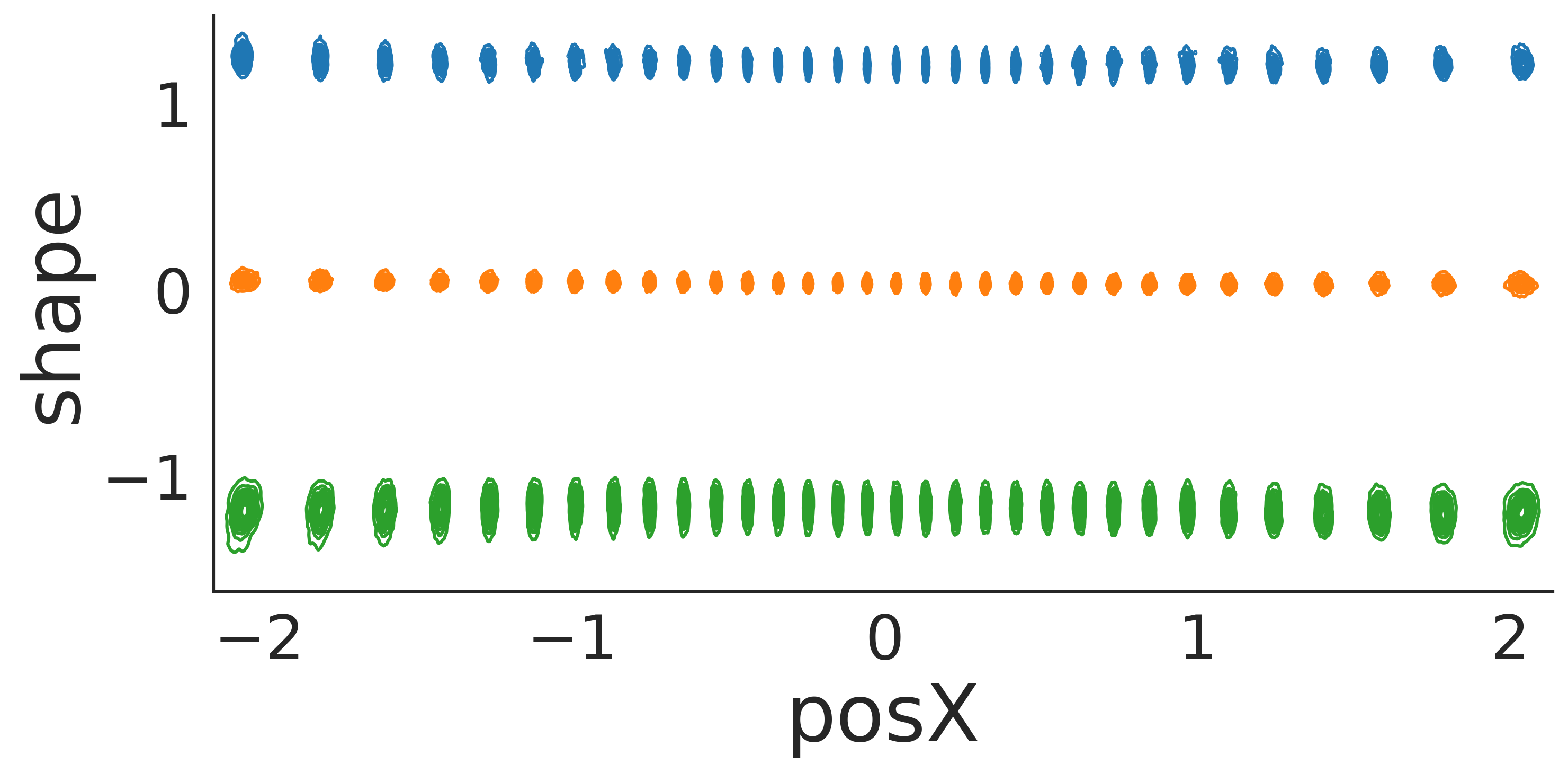}
        \caption{}
        \label{fig:frankenstein-base}
    \end{subfigure}%
    \hspace{1em}
    \begin{subfigure}{\textwidth}
        \centering
        \includegraphics[width=0.7\linewidth]{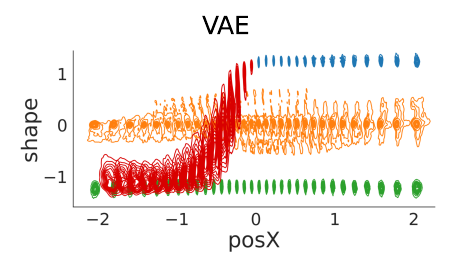}
        \caption{}
        \label{fig:frankenstein-latents}
    \end{subfigure}%
    \hspace{1em}
    \begin{subfigure}{\textwidth}
        \centering
        \includegraphics[width=0.7\linewidth]{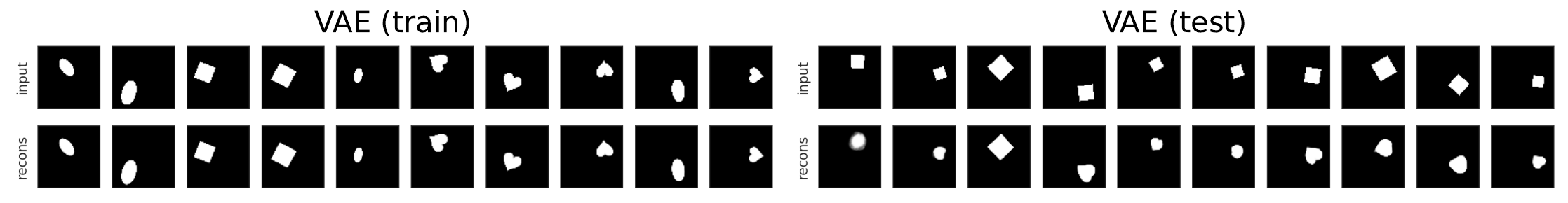}
        \caption{}
        \label{fig:frankenstein-recons}
    \end{subfigure}%
    \caption{\textbf{Encoder trained with a perfect decoder}. We take a model that has achieved high level of disentanglement on the full dataset (top row) and freeze it's docoder. A new encoder is then attached to this perfect decoder and trained on the generalisation case for dSprites producing a set of latent representations (middle) and reconstructions for both training and test data (bottom row).}
\end{figure}

We constructed this ideal decoder (see Figure~\ref{fig:frankenstein-schema}) by training a model in the usual way on the composition task, except the decoder was trained on the entire dataset (with no combinations left out). Once this model had been trained, we verified that the decoder had learned disentangled representations. Figure~\ref{fig:frankenstein-base} visualises the latent space of this model showing that the model did indeed learn highly disentangled representations. We then took a new (untrained) VAE and replaced the decoder with this ``ideal decoder'' and froze all the weights. In the next step we trained the VAE on the image reconstruction task on the dSprites dataset. Critically, this time we left out some combinations of shape and position (the same combination that was left out in Figure~\ref{fig:results-comp}). We then tested this model in the regular manner on samples from all possible combinations of shapes and positions. The results of the reconstructions are shown in Figure~\ref{fig:frankenstein-recons}. We can see that, just like the model where the encoder and decoder were jointly trained, the model manages to reconstruct the training images but fails in most cases to produce the test images accurately. Furthermore, a visualisation of the latent space (Figure~\ref{fig:frankenstein-latents} shows that, despite being trained with this ideal decoder, the encoder fails to map unseen combinations to the correct region of the latent space. These results verify that the failure of the encoder to project images with novel combinations is not the result of the decoder used to train the model.

\FloatBarrier

\clearpage
\subsection{Testing discrete latent representations}

The CascadeVAE \citep{jeong2019learning} model uses discrete latents which are supposed to learn to represent categories such as shape. 

\subsubsection{Encoder and Decoder architectures}

All models used in the main text use continuous latent variables, which are the standard variables used for VAEs. However, our data contains various ordinal variables. For example, \comb{shape} has a discrete set of values for the dSprites dataset. Some recent research shows that using a set of discrete variables leads to lower reconstruction loss and better disentanglement. To test whether using discrete latent variables also leads to better combinatorial generalisation for shape, we tested the CascadeVAE proposed by \citet{jeong2019learning}. This model uses discrete latent variables alongside continuous variables. To infer the values of the continuous ones, they use the standard approach in VAEs, producing mean and standard deviation values of a Gaussian distribution. For the discrete variables they use a top down iterative procedure which is defined by a nested optimization operation:

\begin{align*}
    \underset{\theta,\phi}{\max} & \Bigg(\underset{d^{(1)},\dots,d^{(n)}}{\max}\sum_{i=1}^{n}\mu^{(i)^T} d^{(1)} - \lambda \sum_{i \neq j} d^{(i)^T} d^{(j)} \Bigg) \\
    & - \beta\sum_{i=1}^n D_{KL}(q_{\phi}(z | x^{(i)} || p(z)) \\
    \text{subject to } & || d^{(i)} || = 1 \text{, } d^{(i)} \in {0, 1}^S \text{, } \forall i, 
\end{align*}

where $\mu^{(i)}$ donotes the vector of the likelihood $\log p_\phi (x^{(i)} | z^{(i)}, e_k)$ evaluated at each discrete value $k \in [S]$.

\begin{figure}[h!]
    \centering 
    \begin{subfigure}{\textwidth}
        \centering
        \includegraphics[width=0.9\linewidth]{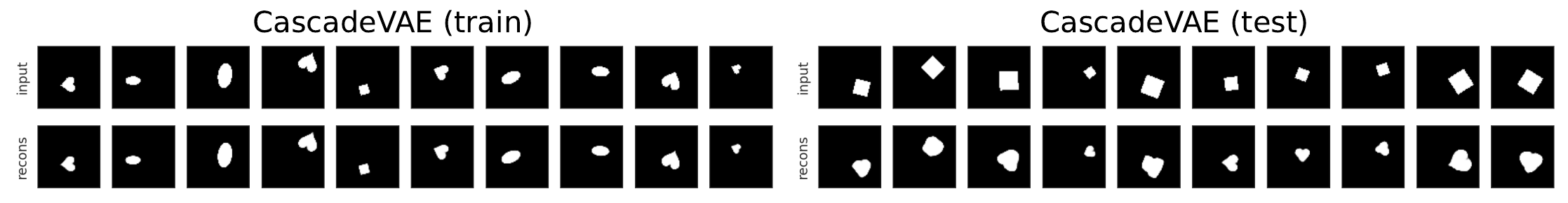}
        \caption{}
    \end{subfigure}%
    \hspace{1em}
    \begin{subfigure}{0.4\textwidth}
        \centering
        \includegraphics[width=0.9\linewidth]{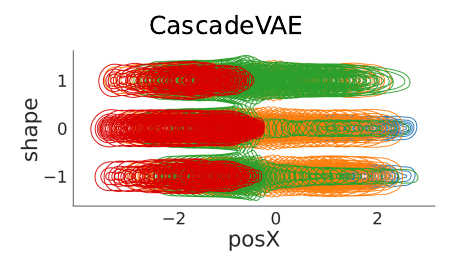}
        \caption{}
    \end{subfigure}%
    \hspace{1em}
    \begin{subfigure}{0.4\textwidth}
        \centering
        \includegraphics[width=0.9\linewidth]{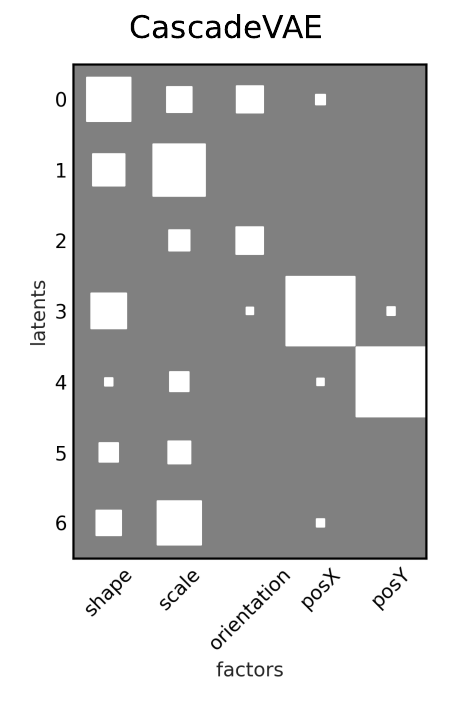}
        \caption{}
    \end{subfigure}%
    \caption{\textbf{Results for CascadeVAE}. (a) shows image reconstructions for training (left) and test (right) for an example model seed. As we can see from these reconstructions, the model successfully trains, reproducing the correct output image for each input image. However, it fails to reconstruct the unseen test combinations in a number of cases, replacing squares with hearts, etc. (b) show the latent space projection for training and test images. We have colour-coded each shape as a different colour and the test cases in red. We can again see that the model fails for the test cases, projecting the unseen shape (square) to three different values of the latent variable. However, it should also be noted here that, unlike the models discussed in the main text, this model does not show a high level of disentanglement, consequently mapping each shape to several values of the latent variable even for the training data. This lack of disentanglement is clear from examining the Hinton matrices in (c).}
    \label{fig:cascade-dsprites}
\end{figure}

The results for the dSprites dataset using CascadeVAE are shown in Figure~\ref{fig:cascade-dsprites}. As can be seen from this figure, the model trains successfully for this dataset, showing impressive reconstruction. However, just like the models tested in the main text, it fails at combinatorial generalisation.

\subsection{Testing interaction between factors}\label{app:lie-group-vae}

\begin{figure}[htbp]
    \centering 
    \begin{subfigure}{\textwidth}
        \centering
        \includegraphics[width=0.8\linewidth]{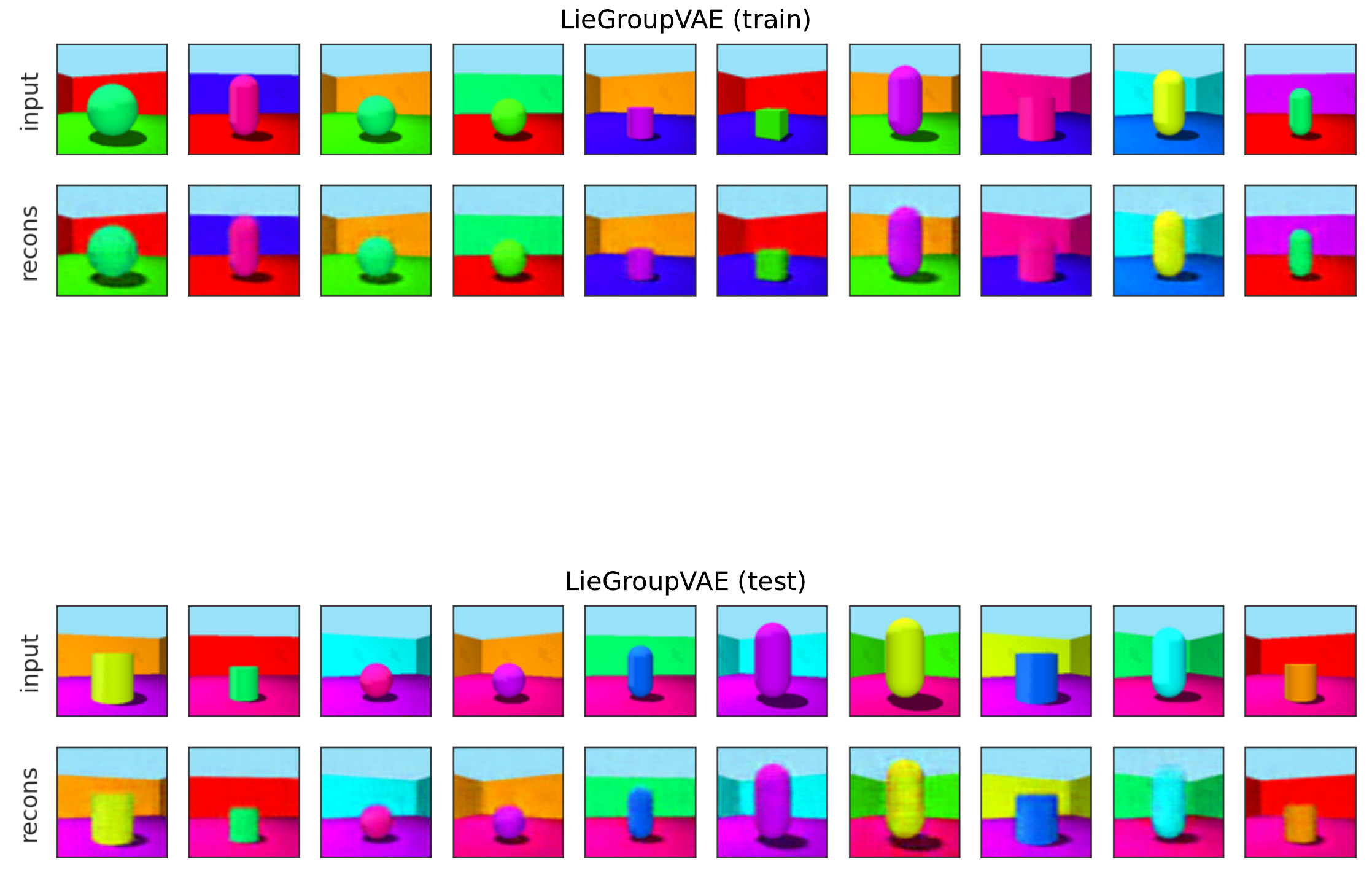}
        \caption{}
    \end{subfigure}%
    \hspace{1em}
    \begin{subfigure}{0.6\textwidth}
        \centering
        \includegraphics[width=0.8\linewidth]{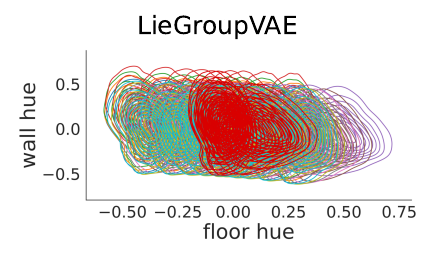}
        \caption{}
    \end{subfigure}%
    \hspace{1em}
    \begin{subfigure}{0.2\textwidth}
        \centering
        \includegraphics[width=0.9\linewidth]{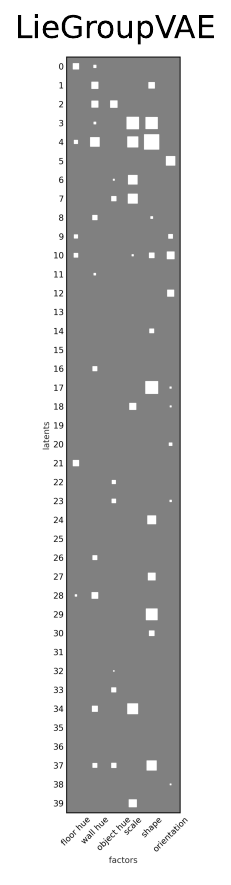}
        \caption{}
    \end{subfigure}%
    \caption{\textbf{Results for LieGroupVAE -- non-interactive condition (3DShapes)}. (a) shows image reconstructions for training (top) and test (bottom) for an example model seed. As we can see from these reconstructions, the model successfully trains, reproducing the correct output image for each input image. It also succeeds at reconstructing the (unseen) test images in this non-interactive condition (note the combinations of wall-hue and floor-hue that are correctly reproduced). (b) show a visualisation of latent space. We have colour-coded each shape as a different colour and the test cases in red. It should also be noted here that, unlike the models discussed in the main text, the multidimensional nature of the subspaces for each subgroup makes it difficult to visualise these in a satisfactory way. (c) visualises the dependencies between generative factors and latent variables using Hinton matrices. In general, we obtained models that were highly disentangled, with disentanglement score $~ 0.90$. In this case the number of latents is much higher since the group size used in the original work is 400 for this dataset instead of 100 as used in dSprites.}
    \label{fig:lie-non-interactive}
\end{figure}

Given our conclusion that failures to model certain interactions appropriately could be the cause of the failures of previous approaches, we tested a recent model -- Commutative LieGroupVAE \citep{zhu_commutative_2021} -- that uses an adaptive equivariant structure, rather than a fixed vector space, to learn factors of variation in the data. This approach combines explicit modeling group operations plus penalties to the learned basis in order to learn a highly disentangled representation of the input (see the original work for more details). The hope is that because this method is adaptive, it may be able to capture not only the generative factors underlying the data, but also the dependencies between them (see Figure~\ref{fig:interactive}).

\begin{figure}[h!]
    \centering 
    \begin{subfigure}{\textwidth}
        \centering
        \includegraphics[width=0.9\linewidth]{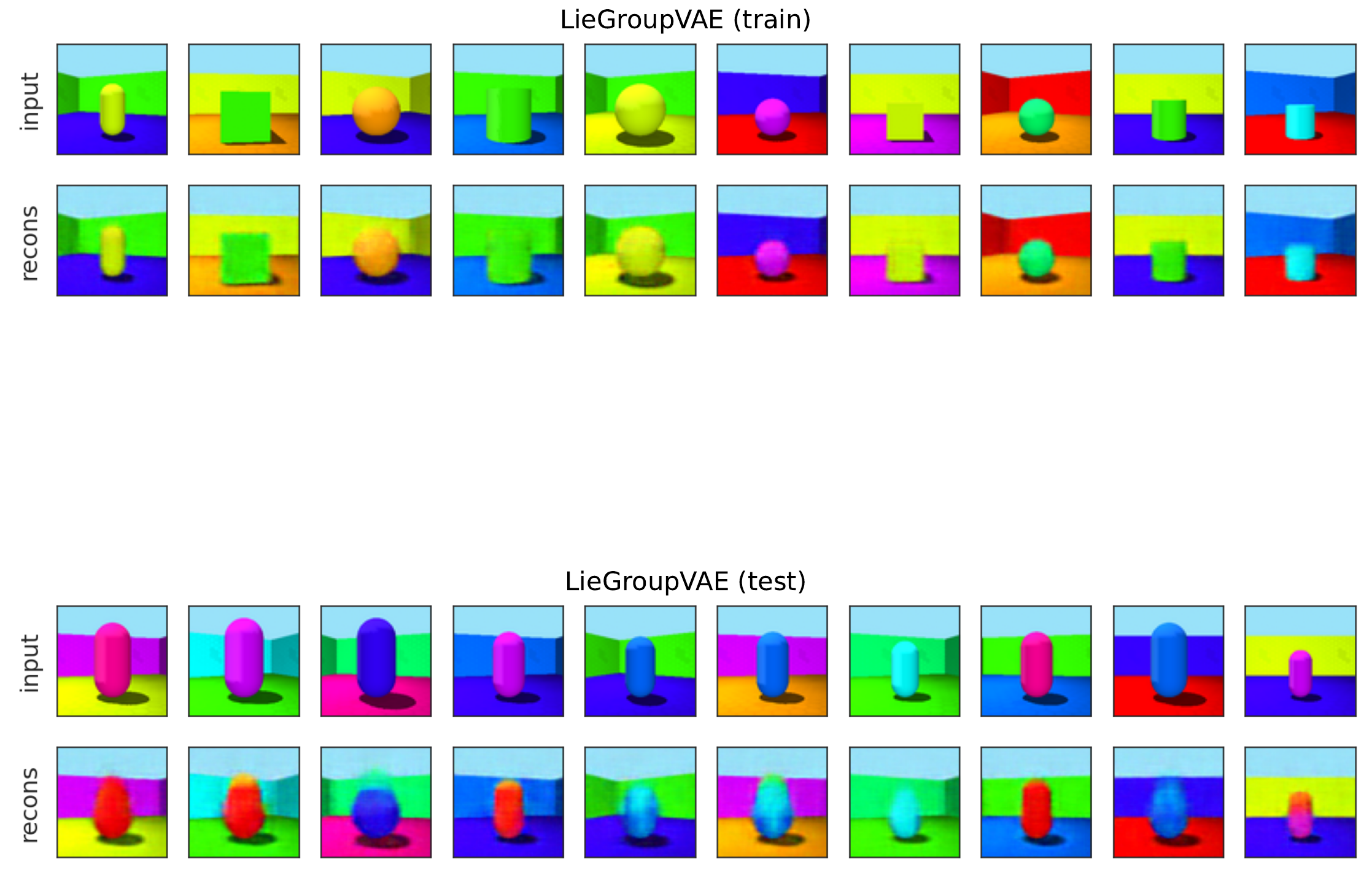}
        \caption{}
    \end{subfigure}%
    \hspace{1em}
    \begin{subfigure}{0.6\textwidth}
        \centering
        \includegraphics[width=0.9\linewidth]{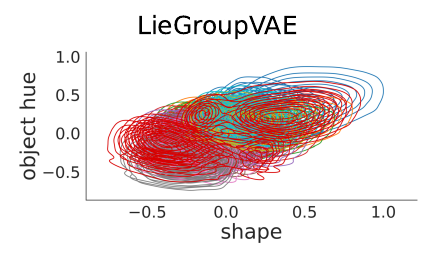}
        \caption{}
    \end{subfigure}%
    \hspace{1em}
    \begin{subfigure}{0.2\textwidth}
        \centering
        \includegraphics[width=0.9\linewidth]{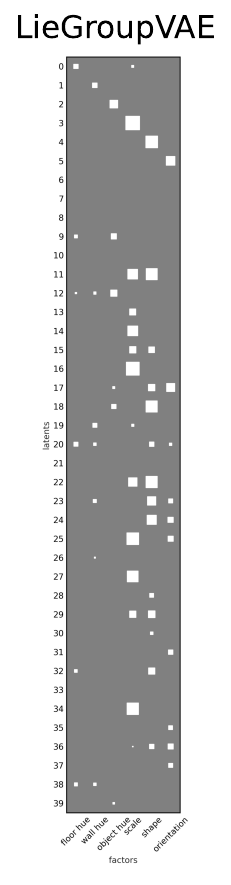}
        \caption{}
    \end{subfigure}%
    \caption{\textbf{Results for LieGroupVAE -- interactive condition (3DShapes)}. (a) shows image reconstructions for training (top) and test (bottom) for an example model seed. Again, the model successfully trains, reproducing the correct output image for each input image. However, in this case, it fails to reconstruct the unseen test combinations in a number of cases, though the mistakes are more varied than with other models, mixing shapes, colours and even combining them in some cases. (b) show the latent space projection for training and test images, where we have colour-coded each shape as a different colour and the test cases are in red. It should also be noted here that, unlike the models discussed in the main text, the multidimensional nature of the subspaces for each subgroup makes it difficult to visualize these in a satisfactory way.}
    \label{fig:lie-interactive}
\end{figure}

To check whether this approach helps overcome the problem of combinatorial generalisation for interactive generative factors, we replicated our experiments with 3DShapes under the interactive and non-interactive conditions. The first simulation tested the condition where the combination \comb{floor hue~$<0.25$, wall hue~$>0.75$} was left out of the training set. This is the non-interactive condition, since the \comb{floor-hue} and \comb{wall-hue} determine different parts of the image. The second simulation tested the condition where the combination \comb{shape=pill, object-hue$>0.5$} was left out of the training set. This is the interactive condition since \comb{shape} and \comb{object-hue} determine the same set of pixels in the image. In addition, we ran a third simulation, on the dSprites dataset, where the combination \comb{shape=square, posX>0.5} was left out of the training set. This is also an interactive condition because shape and position jointly determine the pixels of the image.

Visualizing the latent space of this model is harder, since each factor is now represented in a subspace that is not one-dimensional, but ten-dimensional (we used the original code and parameters from the author's original repository). To plot latent representations, we first projected each subspace to a one-dimensional representation using PCA, and computed disentanglment scores on the result. Disentanglement scores for the 3DShapes simulations reached 0.9 on the interactive case. For each case we ran 5 seeds and have shown the best models in terms of disentanglement (they all achieved similar reconstruction scores).

\begin{figure}[h!]
    \centering 
    \begin{subfigure}{\textwidth}
        \centering
        \includegraphics[width=0.9\linewidth]{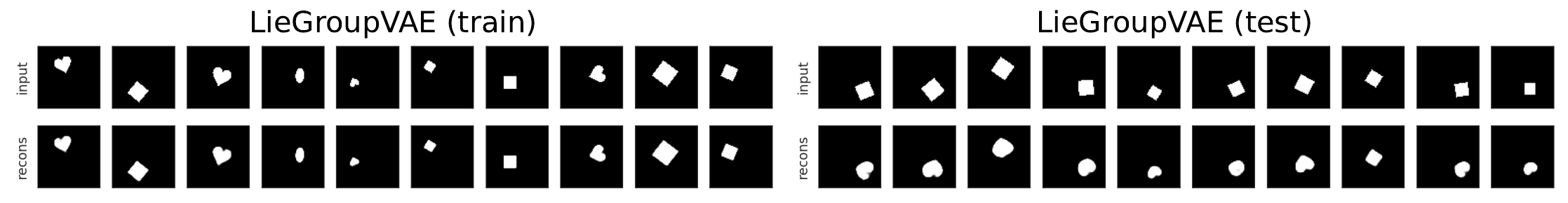}
        \caption{}
    \end{subfigure}%
    \hspace{1em}
    \begin{subfigure}{0.5\textwidth}
        \centering
        \includegraphics[width=0.9\linewidth]{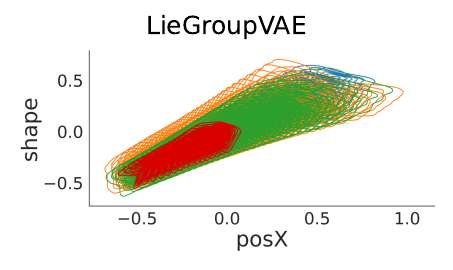}
        \caption{}
    \end{subfigure}%
    \hspace{1em}
    \begin{subfigure}{0.3\textwidth}
        \centering
        \includegraphics[width=0.9\linewidth]{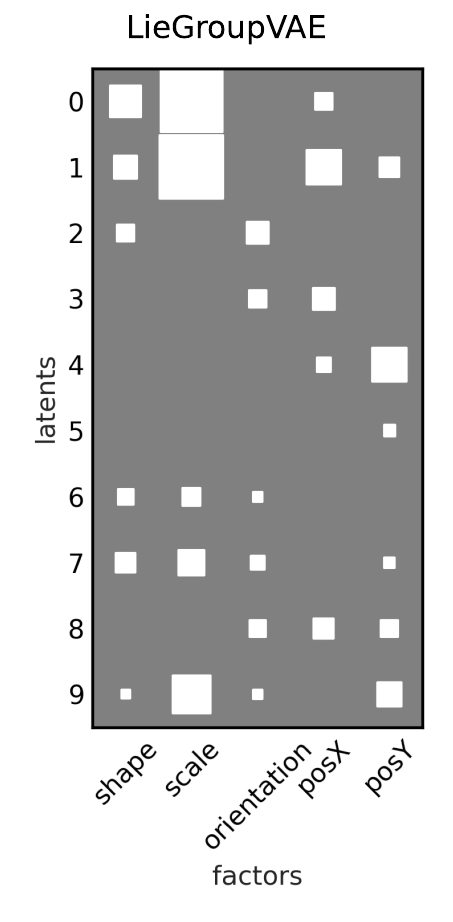}
        \caption{}
    \end{subfigure}%
    \caption{\textbf{Results for LieGroupVAE on dSprites}. (a) shows image reconstructions for training (top) and test (bottom) for an example model seed. Like in the case of 3DShapes (interactive condition), the model successfully trains, reproducing the correct output image for each input image. However, it fails to reconstruct the unseen test combinations in a number of cases, with squares being reconstructed as blobs that resemble hearts (b) show the latent space projection for training and test images. We have colour-coded each shape as a different colour and the test cases in red. (c) shows the Hinton matrices of these models. In all cases, models managed to achieve a disentanglement score $>0.9$}
    \label{fig:lie-dsprites}
\end{figure}

The results of these simulations are presented in Figures~\ref{fig:lie-non-interactive}, \ref{fig:lie-interactive} and~\ref{fig:lie-dsprites}. First of all, like the other models tested in the main manuscript, we observed that the LieGroupVAE manages to learn these tasks and performs image reconstruction to an impressive degree on the training set. Secondly, we found that this model succeeds on the non-interactive condition (\comb{floor hue~$<0.25$, wall hue~$>0.75$}), reconstructing images with unseen combinations of floor-hue and wall-hue successfully (see Figure~\ref{fig:lie-non-interactive}). Lastly, we found that, just like the other models discussed in the manuscript, LieGroupVAE also failed to reconstruct images in the interactive conditions -- \comb{shape=pill, object-hue=$>0.5$} in 3DShapes (Figure~\ref{fig:lie-interactive}) and \comb{shape=square, posX>0.5} in dSprites (Figure~\ref{fig:lie-dsprites}). These simulations suggest that the approach used in LieGroupVAE is not sufficient to overcome the problem of combinatorial generalisation for interactive generative factors. 

\end{document}